%% file: main.tex
\documentclass{article}

\PassOptionsToPackage{numbers,compress}{natbib}
\PassOptionsToPackage{table,xcdraw}{xcolor}

\usepackage[main, final]{neurips_2026}
\usepackage[utf8]{inputenc}
\usepackage[T1]{fontenc}
\usepackage{hyperref}
\usepackage{url}
\usepackage{booktabs}
\usepackage{amsfonts}
\usepackage{nicefrac}
\usepackage{microtype}
\usepackage{booktabs}
\usepackage[table]{xcolor}
\usepackage{graphicx}

\definecolor{headergray}{RGB}{230,230,230}
\definecolor{pastelgreen}{RGB}{190,235,190}

\usepackage{graphicx}
\usepackage{amsmath}
\usepackage{amssymb}
\usepackage{array}
\usepackage{multirow}
\usepackage{float}
\usepackage{siunitx}
\usepackage{algorithm}
\usepackage{algorithmic}
\usepackage{makecell}
\usepackage{capt-of}
\usepackage{subcaption}
\usepackage{xspace}
\usepackage[capitalize]{cleveref}
\usepackage{orcidlink}

\newcommand{\eg}{\emph{e.g.}\xspace}

\input{preamble}

\input{definitions}

\title{FlashControl: One-Step Controllable Generator via
Distillation-Friendly Single-Stream Teachers}

\begin{document}

\maketitle
\begin{center}
\vspace{-25mm}

\begin{tabular}[t]{c}
    \textbf{Ngan Nguyen$^{1*}$, Dung Nguyen$^{1*}$, Quan Dao$^{2}$,} \\
    \textbf{Dimitris N. Metaxas$^{2}$, Minh Hoai$^{1,4}$,
    Cuong Pham$^{1,3}$, Anh Tran$^{1}$} \\[4pt]
    $^{1}$Qualcomm AI Research \\
    $^{2}$Rutgers University \\
    $^{3}$Posts and Telecommunications Institute of Technology \\[4pt]
    $^{4}$Adelaide University \\[4pt]
    \texttt{\{ngannguy,dungtien,minhhoai,pcuong,anhtra\}@qti.qualcomm.com} \\
    \texttt{quan.dao@rutgers.edu, dnm@cs.rutgers.edu}
\end{tabular}
\end{center}
\begin{figure}[h]
\vspace{-5mm}

  \centering
  \includegraphics[width=1\linewidth]{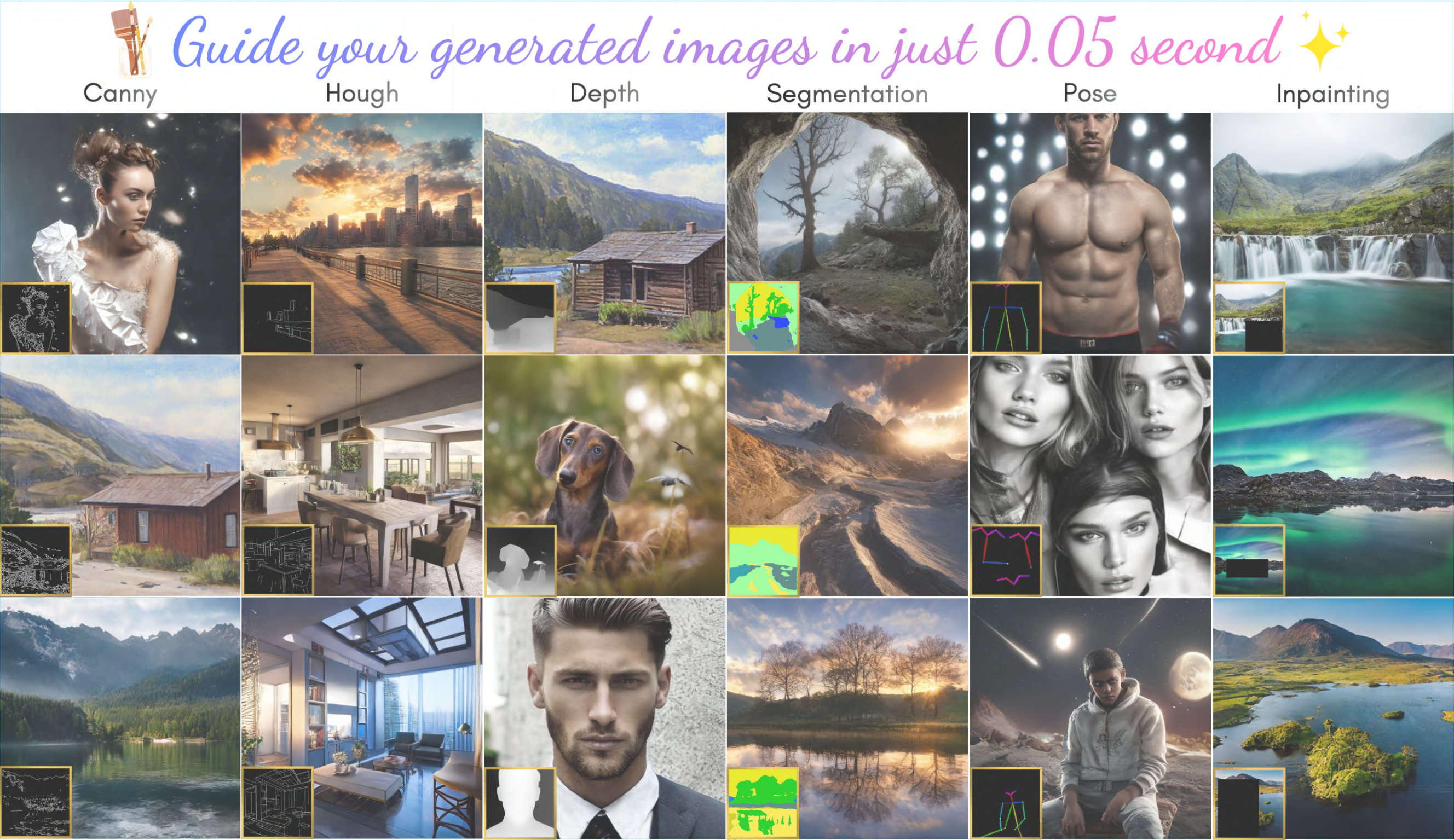}
  \caption{Our controllable image generator, \textbf{FlashControl}, turns spatial conditions into high-quality images in just \textbf{a single step.}}
  \label{fig:teaser}
\end{figure}
\vspace{-5mm}

\input{sec/0_abstract}

\input{sec/1_intro}

\input{sec/2_related_work}
\input{sec/3_background}
\input{sec/4_method}

\input{sec/5_exps}
\input{sec/6_conclusion}

\begin{ack}
% Acknowledgements go here (omitted for anonymous submission).
This work was supported by Qualcomm AI Research.
The authors declare no additional competing interests
beyond their disclosed affiliations.
\end{ack}

\bibliographystyle{abbrvnat}
\bibliography{main}

\appendix
\input{sec/X_suppl}

% ---------------------------------------------------------------

\newpage
\input{checklist}

\end{document}

%% file: preamble.tex
\newcommand{\red}[1]{{\color{red}#1}}

\usepackage{amsmath, amssymb}

%% file: definitions.tex
\def\1n{\mathbf{1}_n}
\def\0{\mathbf{0}}
\def\1{\mathbf{1}}

\definecolor{pink}{rgb}{0.9,0.5,0.5}
\definecolor{purple}{rgb}{0.5, 0.4, 0.8}   
\definecolor{gray}{rgb}{0.3, 0.3, 0.3}
\definecolor{mygreen}{rgb}{0.2, 0.6, 0.2}

\newcommand{\green}[1]{\textcolor{green}{#1}} 

\definecolor{greena}{rgb}{0.4, 0.5, 0.1}

\definecolor{bluea}{rgb}{0, 0.4, 0.6}

\definecolor{reda}{rgb}{0.6, 0.2, 0.1}

\newcommand{\myheading}[1]{\vspace{0.5ex}\noindent \textbf{#1}}

%% file: sec/0_abstract.tex
\begin{abstract}
Diffusion models have become a leading approach in visual generative modeling, particularly effective in text-to-image and controllable generation. Their main drawback is slow inference as it requires many sampling steps. While recent distillation methods enable one-step text-to-image generation, extending these techniques to controllable generation is still challenging. Existing controllable models, such as ControlNet, rely on dual-branch design that make distillation difficult.
In this work, we adopt a simpler strategy. Instead of adding extra conditional branches, we finetune a selected subset of layers in the original text-to-image diffusion model to directly enable controllable generation. This results in \textbf{DeltaControl}, a single-flow, multi-step diffusion model capable of supporting spatial-condition generation while matching the performance of dual-branch approaches. Building on this teacher model, we further apply step-distillation to obtain \textbf{FlashControl}, a one-step model capable of spatial controllable generation with a single forward pass.
\end{abstract}

%% file: sec/1_intro.tex
\section{Introduction}
\label{sec:intro}
Diffusion models \cite{rombach2022high,chen2023pixartalpha,flux2024,ho2020denoising,song2022denoisingdiffusionimplicitmodels} have become a strong paradigm for both text-to-image generation and controllable synthesis. In parallel, recent distillation methods \cite{dao2024swiftbrushv2makeonestep,yin2024improved,yin2024onestep,meng2023distillationguideddiffusionmodels,sauer2023adversarialdiffusiondistillation,salimans2022progressivedistillationfastsampling,nguyen2025superchargedonesteptexttoimagediffusion} have reduced diffusion sampling to a small number of steps, and one-step generators are now increasingly competitive with their multi-step teachers. Extending this progress to controllable generation, however, is substantially less direct. Existing controllable models such as ControlNet and T2I-Adapter inject spatial guidance through external control pathways on top of a pretrained text-to-image backbone~\cite{zhang2023adding,mou2023t2i}. These designs are effective in the standard multi-step setting, but our analysis suggests that they are poorly matched to one-step distillation. When such a teacher is compressed into a single forward-pass student, the student must absorb both the base generative process and the control-specific correction into one computation. In practice, this leads to weak control, unstable optimization, or blurry results, even when the multi-step teacher itself performs well.

This observation motivates the first hypothesis of our work: \emph{for one-step controllable generation to succeed, control must be integrated into the teacher backbone in a form that remains transferable under distillation, rather than attached through an auxiliary branch}. Based on this view, we introduce \textbf{DeltaControl}, a single-stream controllable teacher designed explicitly for one-step transfer. Instead of duplicating large portions of the network, DeltaControl adapts a selected subset of parameters in the original backbone to encode spatial control. We study this design on both UNet and DiT backbones and find that tuning feed-forward layers provides the most favorable behavior for controllability and convergence under a matched parameter budget. Our interpretation is straightforward: in a single-stream architecture, attention layers largely govern global feature mixing, so perturbing them can induce broad and unstable changes, whereas feed-forward layers act on already mixed representations and provide a more stable subspace for injecting control. This yields a controllable teacher that is compact, single-stream, and compatible with standard distillation. More importantly, it distinguishes our goal from prior single-stream controllers such as RepControlNet \cite{deng2024repcontrolnetcontrolnetreparameterization} and OminiControl \cite{tan2025ominicontrol}: our objective is not only efficient controllable generation, but a \emph{distillation-friendly teacher architecture} whose control mechanism survives compression into a one-step student.

Building such a teacher is only the first step. The next challenge is to distill it into a one-step generator without obscuring the effect of teacher architecture. Recent methods such as SwiftBrush \cite{nguyen2025superchargedonesteptexttoimagediffusion,dao2024swiftbrushv2makeonestep} and DMD \cite{yin2024improved,yin2024onestep} show that score-based distribution matching can transfer a multi-step diffusion model to a one-step generator efficiently \cite{yin2024onestep,dao2024swiftbrushv2makeonestep}. In this work, we adopt \textbf{vanilla DMD} \cite{yin2024onestep} as the main distillation objective because it is a simple and widely used score-based formulation that cleanly isolates the role of the teacher architecture. We leave stronger adversarial variants, such as DMD2 \cite{yin2024improved}, to the appendix. This choice is deliberate: our main claim is that teacher design is the critical ingredient for one-step controllable distillation, and we therefore avoid introducing extra adversarial components in the main setting that could confound this comparison.

Even with a suitable teacher and a simple distillation objective, one-step controllable students are still fragile. Different controls, such as canny edges, depth maps, segmentation masks, and poses, have very different structures and scales. If we inject them naively, they can disturb the backbone features and make training unstable. We address this with \textbf{AutoAdapter}. AutoAdapter first aligns condition features to the hidden-state statistics, then applies token-wise gating to control how strongly each token is injected. In our framework, AutoAdapter is not just a small add-on. It is the module that makes different control types usable within one model and keeps early training stable, when the student is most sensitive to feature mismatch.

A separate issue appears during one-step distillation itself. During late-stage distillation, the student starts ignoring variations in its input Gaussian noise, causing different noise samples under the same prompt-condition pair to produce nearly identical outputs. This hurts both diversity and fine detail. To address this, we introduce a \textbf{Latent Variance Preservation} objective, which encourages the output to retain a meaningful response to perturbations in the input noise. The role of this term is specific to one-step diffusion distillation: it preserves stochastic sensitivity in the student and reduces the tendency toward deterministic collapse. Combined with pure DMD and AutoAdapter, it gives a training recipe that is stable enough for one-step controllable generation.

Together, these components form a unified view of the problem. DeltaControl addresses the architectural mismatch between controllable teachers and one-step students. Single stream teacher with pure DMD loss keeps the distillation objective tied to the spatial condition. AutoAdapter aligns heterogeneous controls to the backbone feature space. Latent Variance Preservation keeps the one-step student responsive to noise. Under this formulation, FlashControl is not a loose collection of existing modules. It is a pipeline in which each component resolves a specific failure mode that appears when controllable generation is compressed into a single forward pass.

Our contributions are as follows:
\begin{itemize}
    \item We show that naive one-step distillation fails for dual-branch controllable models, while single-flow teachers provide a better foundation for one-step controllable generation.
    
    \item We introduce \textbf{DeltaControl}, a single-stream controllable teacher that integrates structural control into the base backbone by adapting a selected subset of layers. Across both UNet and DiT backbones, we find that feed-forward tuning provides a favorable balance between controllability, efficiency, and distillation compatibility.    
    \item We show that a DMD-based distillation objective, together with \textbf{AutoAdapter}-based condition alignment, provides a stable recipe for one-step controllable generation and preserves spatial faithfulness under structural controls.
    \item We introduce a \textbf{Latent Variance Preservation} objective for one-step controllable distillation, which reduces deterministic collapse by maintaining the student's sensitivity to input noise, and we demonstrate real-time one-step controllable synthesis across multiple control modalities within a single model.
    \item To the best of our knowledge, \textbf{FlashControl} is among the first one-step controllable distillation frameworks to explicitly build the controllable teacher as a LoRA-compatible single-stream backbone adaptation, rather than distilling from a duplicated control branch.

\end{itemize}

%%%%%%%%%%% END NEW INTRODUCTION %%%%%%%%%%%

%% file: sec/2_related_work.tex
\section{Related Work}
\label{sec:related}

%-------------------------------------------------------------------------
%\anh{Should add a subsection on Diffusion models. I suggest 3 subsections: (1) Diffusion models, (2) Diffusion Distillation, (3) Controllable Diffusion Models.} \ngan{fixed}
\myheading{Diffusion Models} \cite{ho2020denoising, rombach2022highresolutionimagesynthesislatent} have surpassed GAN \cite{goodfellow2014generativeadversarialnetworks} across a wide range of tasks, including controllable generation \cite{zhang2023adding, mou2023t2i}, image editing \cite{Avrahami_2022, meng2022sdeditguidedimagesynthesis}, and personalized synthesis \cite{ruiz2023dreamboothfinetuningtexttoimage}. This progress began with LDM \cite{rombach2022highresolutionimagesynthesislatent} and was further advanced by the DiT architecture \cite{peebles2023scalablediffusionmodelstransformers}. The recent text-to-image models, SD3 \cite{esser2024scalingrectifiedflowtransformers} and FLUX \cite{flux2024, labs2025flux1kontextflowmatching}, achieve impressive results by adopting Rectified Flow training \cite{liu2022flowstraightfastlearning}, and Multi-Modal Transformer (MMDiT) backbone \cite{peebles2023scalablediffusionmodelstransformers}.

\myheading{Multi-step Controllable Diffusion Models.}
Controllable synthesis guides text-to-image diffusion with structural inputs (e.g., edges, depth, segmentation, pose). T2I-Adapter \cite{mou2023t2i} trains lightweight adapters on a frozen backbone, while ControlNet \cite{zhang2023adding} adds a parallel control branch and has inspired many variants, including ControlNet-XS \cite{zavadski2024controlnet}, UniControlNet \cite{zhao2023uni}, UniControl \cite{qin2023unicontrol}, and AnyControl \cite{sun2024anycontrol}. To reduce tuning cost, CTRLORA \cite{xu2024ctrlora} adapts small parameter subsets of the control branch. Instead of using dual-branch, some recent works, such as UniCombine \cite{wang2025unicombine} and OminiControl \cite{tan2025ominicontrol}, append the conditional inputs to the noisy latent input and denoise them with a LoRA-finetuned single-branch DiT. These techniques often leverage strong backbones like FLUX and incur a massive computation cost growth with respect to the number of conditions.
% ecent works such as UniCombine \cite{wang2025unicombine} and OminiControl \cite{tan2025ominicontrol} leverage strong backbones like FLUX. 
Despite strong controllability, these methods still rely on multi-step sampling, leading to slow inference.
%\anh{Too long. Group and split it for single-control and multi-control techniques. Remove a few unimportant works, or group multiple works to say at once.} \anh{Discuss their common speed issue due to multi-step diffusion} \ngan{fixed}

% \quan{related follows history: controlnet (single-control, multistep) -> efficient controlnet (ctrllora) -> mult-control, multistep. Distill controlnet. Aim multicontrol, singlestep. Distill controlnet is hard due to dual flow -> propose single-flow technique (simple finetune FFN) which is compatiable with technique distillation }
%-------------------------------------------------------------------------
\myheading{Diffusion Distillation.}
Diffusion models are slow at inference due to many denoising steps, motivating distillation to shorten sampling while preserving quality \cite{salimans2022progressivedistillationfastsampling, lu2022dpmsolverfastodesolver}. 
Consistency Models and LCM enable few-step sampling but still show a quality gap at one step \cite{song2023consistencymodels, luo2023latentconsistencymodelssynthesizing}. 
ADD/LADD use adversarial losses for 1-4 step distillation but can be unstable \cite{sauer2023adversarialdiffusiondistillation, sauer2024fasthighresolutionimagesynthesis}, while DMD/DMD2 distill from the teacher score field for one-/few-step generation \cite{yin2024onestep,yin2024improved}. SwiftBrush reduces overhead via image-free score distillation with VSD \cite{thuan2024SB, wang2024prolificdreamer}, and SBv2 improves further with stronger initialization, efficient LoRA, and a clamped-margin CLIP loss, sometimes surpassing the teacher \cite{dao2024swiftbrushv2makeonestep}.

%% file: sec/3_background.tex
\section{Preliminaries}
%This section introduces the foundational concepts underlying our work, including diffusion models, ControlNet, and DMD technique.

\begin{figure}
  \centering
  \includegraphics[width=0.8\linewidth]{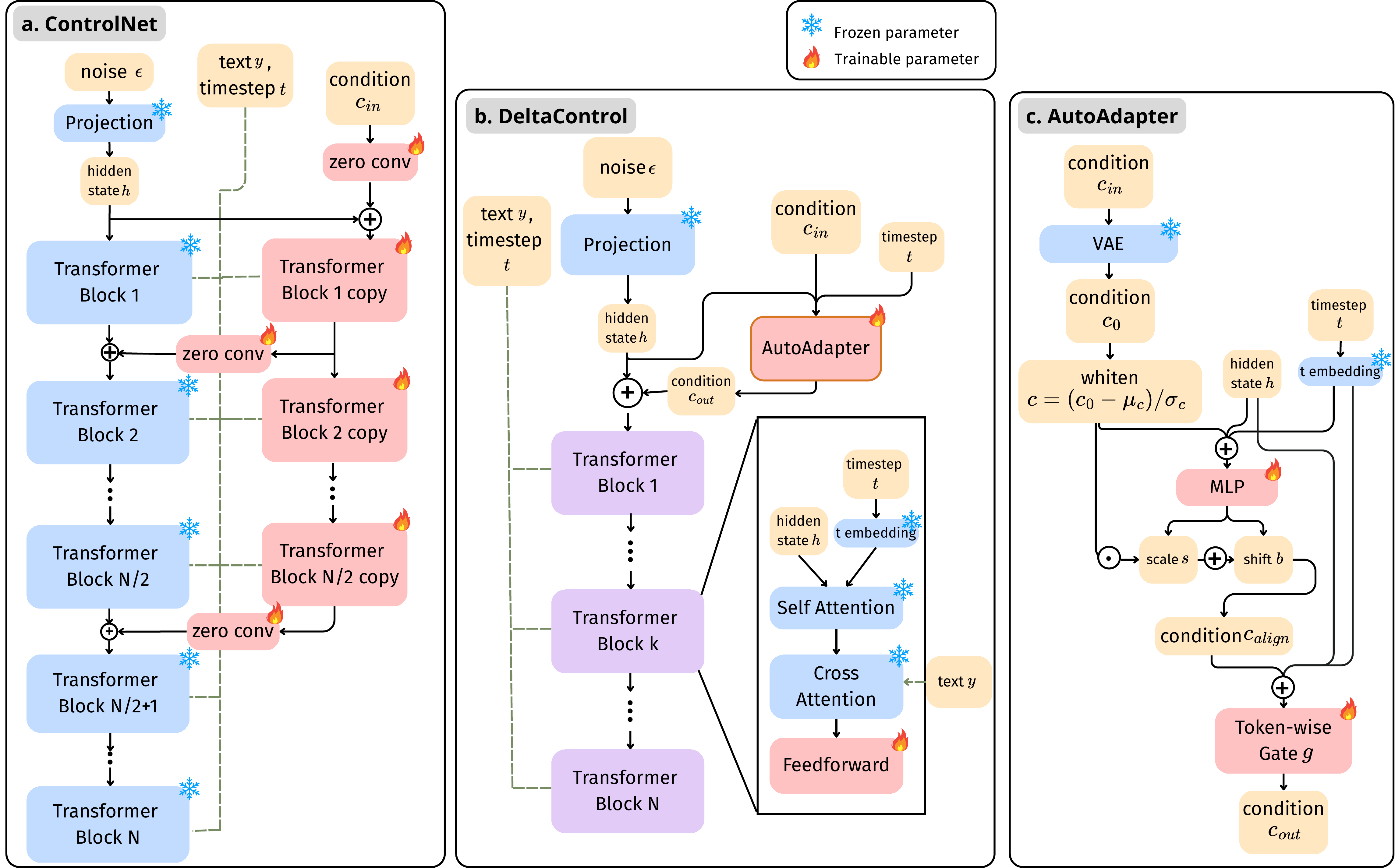}
  % \vspace{-6mm}
  \caption{Comparison of ControlNet (a) and our proposal \textbf{DeltaControl} (b). The architecture of AutoAdapter is illustrated in (c).}
  \label{fig:archi}
  \vspace{-5mm}
\end{figure}

\paragraph{Diffusion and Flow-based Generative Models.}
Modern text-to-image models generate images by learning to map a noisy sample back to clean data. 
Let $x_0$ denote a clean image or latent, $t \in [0,1]$ a timestep, and $\epsilon \sim \mathcal{N}(0,I)$. 
A common noising process is
\[
x_t = \alpha_t x_0 + \sigma_t \epsilon ,
\]
where $\alpha_t$ and $\sigma_t$ define the signal and noise scales. 
Depending on the model family, the denoiser predicts either the noise $\epsilon$ or a velocity/flow target $v_t$. 
We write this generically as
\[
f_\phi : (x_t,t,y) \mapsto \hat{r}_t,
\qquad
\mathcal{L}_{\mathrm{gen}}
=
\mathbb{E}_{x_0,y,t,\epsilon}
\left[
\|f_\phi(x_t,t,y)-r_t\|_2^2
\right],
\]
where $r_t=\epsilon$ for noise-prediction diffusion models and $r_t=v_t$ for velocity or flow-based models. 
At inference, sampling starts from Gaussian noise and repeatedly applies the learned denoising or flow field, which yields high quality but requires many network evaluations.

\paragraph{ControlNet.}
Text prompts provide global semantic guidance, but many applications require spatial control from a condition map $c$, such as edges, depth, segmentation, or pose. 
ControlNet~\cite{zhang2023adding} addresses this by attaching an auxiliary control branch to a pretrained text-to-image backbone. 
The control branch processes $(x_t,t,c)$ and injects condition-dependent residuals into the frozen base denoiser, usually through zero-initialized layers. 
This design preserves the pretrained generative capability while learning spatial alignment. 
%However, it also decomposes controllable generation into two coupled paths: the base denoising path and the control-residual path. 
% This decomposition is effective for multi-step sampling, but creates an architectural mismatch for one-step distillation, where a single student evaluation must absorb both paths at once.

\paragraph{Distribution Matching Distillation.}
We use Distribution Matching Distillation (DMD)~\cite{yin2024onestep,yin2024improved} as our main one-step distillation objective. 
DMD trains a generator $G_\theta$ to match the distribution of a pretrained diffusion teacher without simulating the full teacher trajectory at inference. 
Given input noise $z$, text prompt $y$, and spatial condition $c$, the student predicts
\[
\hat{x}_0 = G_\theta(z,y,c).
\]
The student output is then noised to timestep $t$ to obtain $x_t$. 
DMD updates the student using the score discrepancy between a frozen teacher score estimator $\epsilon_\phi$ and a trainable fake score estimator $\epsilon_\psi$:
\[
\nabla_\theta \mathcal{L}_{\mathrm{DMD}}
=
\mathbb{E}_{z,\epsilon,t,y,c}
\left[
w(t)
\left(
\epsilon_\phi(x_t,t,y,c)
-
\epsilon_\psi(x_t,t,y,c)
\right)
\frac{\partial G_\theta(z,y,c)}{\partial \theta}
\right],
\]
where $w(t)$ is a timestep-dependent weighting function. 

%% file: sec/4_method.tex
\newcommand{\ngan}[1]{\textcolor{purple}{(Ngan: #1)}}
\newcommand{\quan}[1]{\textcolor{blue}{(Quan: #1)}}
\newcommand{\dung}[1]{\textcolor{gray}(Dung: #1}

\section{FlashControl}

\subsection{Overview}
% Our objective is \textbf{one-step (single forward pass) text-to-image generation via distillation}, with \textbf{stable training} and \textbf{high-quality multi-condition control}. We hypothesize that \textbf{one-step controllable generation is substantially harder than one-step text-only generation}, and the common dual-branch architecture ControlNet is not distillation-friendly in one-step setting, which is discussed later in \cref{41}. To address this problem, we propose \textbf{FlashControl}, a two-stage pipeline (teacher $\rightarrow$ student), which is real-time, high-quality multi-controllable.

Our objective is \textbf{one-step (single forward pass) controllable text-to-image generation via distillation}, with \textbf{stable optimization} and \textbf{high-quality spatial condition model}. We hypothesize that \textbf{one-step controllable generation is substantially harder than one-step text-only generation}, because the student must preserve both visual fidelity and spatial faithfulness within a single network evaluation. In particular, the common dual-branch design used by ControlNet-style methods is not well matched to one-step distillation, as discussed later in \cref{41}. To address this problem, we propose \textbf{FlashControl}, a two-stage teacher$\rightarrow$student pipeline for real-time, high-quality and controllable generation.

\textbf{Stage 1: distillation-friendly conditional teacher (DeltaControl in \cref{42}).
}
One-step text-to-image distillation methods typically rely on a single-branch teacher architecture. Motivated by this observation, we train a single-branch DeltaControl teacher, which enables spatial signal control by fine-tuning all feedforward layers from the base text-to-image diffusion model. 
%Compared with dual-branch architectures such as ControlNet, DeltaControl is more efficient in both memory usage and training, while maintaining a single-flow controllable teacher that is compatible with existing distillation techniques.
To better handle heterogeneous control modalities, we introduce \textbf{AutoAdapter}, a lightweight conditioning module. 
%that (i) aligns control embeddings to the backbone hidden-state statistics and (ii) applies token-wise reliability gating based on compatibility and uncertainty. This improves DeltaControl in Stage 1 and, crucially, provides a more stable control interface for Stage 2.

\textbf{Stage 2: one-step student (FlashControl) via distillation in \cref{43}.}
We distill the single-stream DeltaControl teacher into a one-step generator using \textbf{Distribution Matching Distillation (DMD)}~\cite{yin2024onestep,yin2024improved}. %Specifically, the student is optimized using the discrepancy between a frozen teacher score estimator and a trainable fake score estimator, allowing the one-step model to match the teacher distribution without simulating a full multi-step trajectory at inference time. 
%In this stage, we retain AutoAdapter unchanged so that the student inherits the same normalized, reliability-weighted control interface learned by the teacher. 
In addition, we deliberately \textbf{drop adversarial distillation}, since controllable generation requires realism under a spatial constraint rather than unconstrained image realism. Finally, to prevent late-stage collapse in one-step distillation, we introduce \textbf{Latent Variance Preservation (LSP)}, which maintain the output variation and prevents deterministic mapping.

\subsection{From Dual-branch to efficient single-stream teacher.}
\label{41}

\paragraph{Is dual-branch ControlNet optimal for one-step distillation?}
ControlNet \cite{zhang2023adding} is the most common framework for multi-step controllable image generation.
A natural approach for \emph{one-step} controllable T2I is therefore to (i) train a ControlNet teacher and (ii) apply a one-step distillation method (e.g., DMD \cite{yin2024improved, yin2024onestep}). We experiment with distilling dual-branch models, including ControlNet and T2I-Adapter, into onestep generators using DMD.
However, the results show that this straightforward pipeline results in low-quality and blurry results, as in \cref{fig:controlnet_fail} and \cref{tab:teacher_arch_distillation}, even when the multi-step teacher itself performs well. 

We attribute this failure to an architectural mismatch between \emph{dual-branch teacher} and \emph{single-step student}.
ControlNet decomposes controllable generation into two coupled components: a base denoiser and an auxiliary control branch that injects condition-dependent residuals.
In one-step distillation, the student must simultaneously approximate the base denoising dynamics \emph{and} the control-induced correction within a single forward pass.
This entangled objective leads to unstable optimization: the student either collapses to the base model (weak control) or overfits noisy control residuals (degrading fidelity).
This behavior is consistent with recent observations in distilling large dual-branch architectures, where additional stabilization tricks are required and training may still collapse (e.g., gradient explosion) \cite{chen2025sanasprint}.
We further verify in \cref{fig:controlnet_fail} that simply ``simplifying'' the objective by freezing either the base denoiser or the control branch does not resolve the issue.
Overall, these results suggest that ControlNet’s dual-branch architecture is not well suited as a teacher for one-step controllable distillation.

\paragraph{Why DeltaControl? From RepControlNet to a unified single-branch teacher.}
This analysis suggests a simple principle: a strong one-step controllable student should be distilled from a single-branch controllable teacher, where control is integrated into the backbone rather than produced by an external branch.
RepControlNet \cite{deng2024repcontrolnetcontrolnetreparameterization} is a representative approach in this direction. It reparameterizes control by replicating parts of the backbone and learning condition-dependent residual transformations that can be merged into a single network, removing the need for a dual control branch during inference.

We identify several limitations of RepControlNet.
First, it replicates parts of the backbone during training following on a tedious replicate-and-merge strategy, leading to unnecessary increase in memory and computation cost. Second, it does not investigate which layers should be adapted to effectively incorporate control signals. Instead, it adopts a dense “replicate-and-tune” strategy that duplicates and finetunes many layers without analyzing their roles in controllability.
Third, RepControlNet is evaluated only on the UNet backbone, leaving open the question of whether the approach generalizes to transformer-based architectures such as DiT.
Finally, RepControlNet focuses primarily on a single-condition setting and evaluates only a limited set of control modalities. This restricts its applicability in scenarios where multiple heterogeneous control signals must be handled within a unified framework.

In contrast, our method addresses all four limitations. We do not replicate network components but directly finetune some of them, offer more efficient training and memory usage. We perform an analysis to identify which layers are most effective for adapting control signals, design a strategy that generalizes across both UNet and transformer backbones, and develop a unified controllable framework capable of handling diverse control modalities within a single model.

\paragraph{Comparison to other single-branch approaches.} 
% Some recent techniques such as OminiControl \cite{tan2025ominicontrol} and UniCombine \cite{wang2025unicombine} concatenate the VAE encodings of the conditional images to the noisy latent input before passing to the denoising network, offering a cleaner DiT-based single-stream formulation for versatile multi-step control. However, compared to DeltaControl, these approaches still have  these approaches have While these sing-branch networks may be friendly for distillation, which need more investigation to confirm, they substantially increase the computation cost. When adding a single condition, the number of input patches grows $2\times$, and the computation cost in self-attention layers grows $4\times$. In contrast, DeltaControl only introduces a minimal overhead for AutoAdapter encoding. 
% Not all single-stream controllable architectures are equally suitable as teachers for one-step distillation. 
Recent methods, such as OminiControl \cite{tan2025ominicontrol} and UniCombine \cite{wang2025unicombine}, append the conditions to the noisy latent input, offering a cleaner DiT-based single-stream formulation for versatile multi-step control. However, their goal is efficient controllable generation itself, not the construction of a teacher whose control interface can be faithfully compressed into a one-step student. In contrast, DeltaControl is designed around this stricter requirement: it uses \textbf{selective backbone adaptation} to integrate control directly into layers that remain stable under transfer, and combines this with AutoAdapter so that the student inherits a normalized, reliability-aware control interface during distillation. As a result, our single-stream design is not merely lighter; it is more friendly to distill, as will be showed in \Cref{sec:exps}.

\subsection{Stage 1: DeltaControl - Multi-step Conditional Model}\label{42}
%We propose DeltaControl, an efficient controllable framework compatible with standard distillation techniques. It achieves controllability by finetuning a selected subset of parameters in the base model and introduces AutoAdapter to support unified spatial control.

\subsubsection{Efficient Finetuning Strategy.}
Inspired by RepControlNet, DeltaControl removes the dual-branch architecture and instead fine-tunes only a selected subset of parameters in the base T2I model.
To determine which layers are most suitable for learning control signals, we conduct a series of experiments on two widely used diffusion backbones: UNet and DiT. For DiT, which consists of Transformer blocks combining self-attention, cross-attention, and feed-forward (FFN) layers, we evaluate three finetuning configurations with the same number of trainable parameters as vanilla ControlNet: (i) attention-only, (ii) FFN-only, and (iii) a mixed subset of attention and FFN layers. For UNet, which contains both Transformer and ResNet blocks, we examine three analogous variants: (i) ResNet - selecting a subset of ResNet blocks with the same number of trainable parameters as ControlNet, (ii) Attn - replacing part of the selected ResNet blocks with attention layers, and (iii) FFN - replacing part of the ResNet subset with feed-forward layers. We use Stable Diffusion 2.1 \cite{rombach2022high} as the UNet-based model and PixArt-$\alpha$ \cite{chen2023pixartalpha} as the DiT-based model. During training, only the selected subset of parameters is fine-tuned. Control signals are injected using a simple condition embedding, implemented as a learnable MLP, and we evaluate controllability using the Canny edge condition.

As shown in ~\cref{fig:vary_params}, \textbf{finetuning FFN layers consistently yields the largest improvements} in adapting to new conditioning signals across both architectures, compared to finetuning attention layers alone. Notably, this strategy consistently outperforms ControlNet in both UNet and DiT backbones across all training stages. We hypothesize that this effect arises because, in a single-flow architecture, the self- and cross-attention modules already align textual prompts with image features, while the \textbf{FFN layers mainly transform and refine the combined representations}. In contrast, modifying attention layers changes the global mixing patterns of features, where small parameter updates can lead to disproportionately large and potentially destabilizing effects.

We illustrate the architectural differences between ControlNet and DeltaControl in \cref{fig:archi}. Instead of training an additional control branch, DeltaControl achieves controllability by fine-tuning a small subset of parameters in the base model, significantly reducing overhead compared to ControlNet and RepControlNet. As shown in Tab.\ref{tab:train_efficiency}, this design requires fewer trainable parameters, leading to lower GPU memory usage and faster training. At inference time, DeltaControl runs a single-branch that natively incorporates control signals, eliminating the need for a separate control branch. The selected parameters can also be adapted with LoRA, preserving the original model’s generative capability. Overall, this design improves both training and inference efficiency. In addition, we introduce AutoAdapter, which enables effective controllable generation.

%\anh{need a system figure to explain the structure, particularly the Auto Align Adapter} \ngan{I drew it but it seems we are running out of paper } \anh{Please add it first. Will shrink the text later} \ngan{added}

% Removing the ControlNet branch means we must inject control signals into a single stream without disrupting the backbone’s feature statistics. To help the model interpret these signals, we introduce \textbf{AutoAdapter}, a lightweight module for DeltaControl that handles spatial inputs and plugs in at the head of the diffusion model. We present the detail of this module in the next section, note that \cref{fig:vary_params} does not leverage this module.

\subsubsection{AutoAdapter.}

A direct addition of raw control embeddings to the model's hidden states is fragile in the multi-condition setting: edge, depth, segmentation, and pose signals differ substantially in scale, sparsity, and reliability. These differences can cause either vanishing control signals or over-dominant perturbations. To address this, we introduce \textbf{AutoAdapter}, a lightweight module that converts incoming control tokens into hidden-state-compatible residuals prior to backbone injection. Its primary role is twofold: it improves the quality of DeltaControl under heterogeneous conditions and it provides a stable control interface for the alternating student-teacher updates during distillation.

Given a control image $c_{\text{in}}$, we encode it using the base model's VAE encoder to obtain latent control tokens $c_0$. AutoAdapter then sequentially applies a \textit{distribution matching} module and a \textit{token-wise reliability gate}. The distribution matching module aligns the first- and second-order statistics of the control tokens to the backbone hidden-state scale:
\[
c=\frac{c_0-\mu_c}{\sigma_c}, \qquad
s,b=\mathrm{MLP}(h,c,\phi(t)), \qquad
c_{\text{align}}=s\odot c+b .
\]
where $\mu_c$ and $\sigma_c$ denote the per-channel mean and standard deviation of $c_0$, $h$ is the current backbone hidden state, and $\phi(t)$ is the timestep embedding. 

% \ngan{Plot $delta_stat$ here, or sorften the framing.}

AutoAdapter is fundamentally distinct from generic feature projectors. Its explicit objective is to minimize the statistical mismatch between the control stream and the backbone stream before fusion. Formally, we define this mismatch as:
$$\Delta_{\text{stat}}(h, c_{\text{align}}) = \lVert \mu(c_{\text{align}}) - \mu(h) \rVert_2^2 + \lVert \sigma(c_{\text{align}}) - \sigma(h) \rVert_2^2.$$
This whitening and rescaling process is critical because the denoiser is locally highly sensitive to hidden-state perturbations. Under a first-order approximation, if $\delta$ denotes the injected control residual, the perturbed prediction behaves as:
$$\epsilon_\theta(h + \delta, t, y) \approx \epsilon_\theta(h, t, y) + J_\theta(h) \delta,$$
where $J_\theta(h)$ is the Jacobian of the denoiser with respect to the hidden state. When $\delta$ is poorly scaled, the resulting update becomes erratic. By ensuring $\delta$ matches the backbone statistics, the control signal enters a manifold the pretrained model can reliably absorb, facilitating superior control convergence.
Following statistical alignment, AutoAdapter applies a token-wise reliability gate to suppress noisy or incompatible condition tokens. For the $i$-th token, the gate is computed as:
\[
g_i=\sigma\!\left(\alpha\,\cos(h_i,c_i^{\text{align}})
+\beta\,\mathrm{time}(t)
-\gamma\,\mathrm{MLP}(c_i^{\text{align}})\right), \qquad
h_i \leftarrow h_i + g_i\, c_i^{\text{align}} .
\]
where $\sigma$ is the sigmoid function, and $\alpha, \beta, \gamma$ are learnable scalars. The first term assesses compatibility between the current hidden token and the aligned condition token; the second term encourages stronger spatial guidance during early denoising steps when the global layout is formed; the third term penalizes unreliable condition tokens. Because $g_i \in (0, 1)$, the injected residual is strictly bounded:
$$\lVert \delta_i \rVert_2 = \lVert g_i c^{\text{align}}_i \rVert_2 \le \lVert c^{\text{align}}_i \rVert_2.$$
This guarantees that the gate does not fabricate new control content, but merely selectively attenuates the already-aligned control signal.

% END NEW AA

% \[
% g \;=\; \sigma\!\big(W_2\,\mathrm{SiLU}(W_1 z)\big)
% \;\in\; [0,1]^{B\times \min(N,\tilde N)\times 1}.
% \]

% \[
% r \;=\; g \odot c_{\text{align}}, \qquad
% h \;\leftarrow\; h + \mathrm{Proj}(r),
% \]
% where $\mathrm{Proj}(\cdot)$ is a light linear/projection to match token grids or
% channels (a $1{\times}1$ conv or linear). We zero-init $W_2$ so that $g\!\approx\!0$
% at start, guaranteeing safe warm-up.
\begin{table*}[!t]
  \caption{\textbf{Quantitative comparison} across six control tasks regarding to FID, Rec, Prec on COCO-5k. "-": the model does not support that condition.}

  \centering
  \renewcommand{\arraystretch}{1.2}
  \resizebox{\textwidth}{!}{
    \begin{tabular}{l c c c c c c c c c c c c c c c c c c c c}
      \toprule
      \textbf{Model} & \textbf{Backbone} & \textbf{NFEs} & \multicolumn{3}{c}{Canny} & \multicolumn{3}{c}{Depth} & \multicolumn{3}{c}{Seg} & \multicolumn{3}{c}{Pose} & \multicolumn{3}{c}{Hough} & \multicolumn{3}{c}{Inpainting} \\
      \cmidrule(lr){4-6} \cmidrule(lr){7-9} \cmidrule(lr){10-12} \cmidrule(lr){13-15} \cmidrule(lr){16-18} \cmidrule(lr){19-21}
      & & & FID$\downarrow$ & Rec$\uparrow$ & Prec$\uparrow$ & FID$\downarrow$ & Rec$\uparrow$ & Prec$\uparrow$ & FID$\downarrow$ & Rec$\uparrow$ & Prec$\uparrow$ & FID$\downarrow$ & Rec$\uparrow$ & Prec$\uparrow$ & FID$\downarrow$ & Rec$\uparrow$ & Prec$\uparrow$ & FID$\downarrow$ & Rec$\uparrow$ & Prec$\uparrow$ \\
      \midrule

      % ===================== SD1.5 =====================
      ControlNet \cite{zhang2023adding} & SD1.5 & 50 & 19.86 & 0.72 & 0.65 & 30.04 & 0.53 & 0.42 & 29.18 & 0.53 & 0.52 & 28.45 & 0.49 & 0.56 & 30.75 & 0.53 & 0.49 & - & - & - \\
      UniControlNet \cite{zhao2023uni} & SD1.5 & 50 & 17.79 & 0.63 & 0.65 & 21.2 & 0.57 & 0.54 & 23.08 & 0.47 & 0.49 & 27.85 & 0.36 & 0.49 & 31.36 & 0.46 & 0.44 & - & - & - \\
      Uni-Control \cite{qin2023unicontrol} & SD1.5 & 50 & 22.90 & 0.72 & 0.59 & 21.3 & 0.60 & 0.50 & 25.50 & 0.51 & 0.46 & 27.4 & 0.43 & 0.43 & - & - & - & 19.85 & 0.79 & 0.75 \\
      T2I adapter \cite{mou2023t2i} & SD1.5 & 50 & 17.59 & 0.65 & 0.56 & 20.73 & 0.55 & 0.54 & - & - & - & 25.62 & 0.42 & 0.48 & - & - & - & - & - & - \\
      AnyControl \cite{sun2024anycontrol} & SD1.5 & 50 & 18.55 & 0.59 & 0.62 & 19.32 & 0.55 & 0.57 & - & - & - & 24.99 & 0.38 & 0.56 & - & - & - & - & - & - \\
      ControlNet++ \cite{controlnet_plus_plus} & SD1.5 & 50 & 21.01 & 0.757 & 0.726 & 19.18 & 0.718 & 0.641 & 48.15 & 0.438 & 0.478 & - & - & - & - & - & - & - & - & - \\

      CTRLORA \cite{xu2024ctrlora} & SD1.5 & 50 & 16.65 & 0.70 & 0.55 & 19.34 & 0.57 & 0.50 & 21.13 & 0.49 & 0.48 & 25.18 & 0.42 & 0.46 & - & - & - & - & - & - \\
      LCM-ControlNet \cite{chen2024pixartdelta} & SD1.5 & 4 & 25.83 & 0.71 & 0.65 & 28.11 & 0.41 & 0.50 & 31.55 & 0.43 & 0.50 & 37.19 & 0.33 & 0.56 & 38.71 & 0.41 & 0.49 & - & - & - \\
      FlashSD \cite{chadebec2024flash} & SD1.5 & 4 & 20.42 & 0.71 & 0.65 & 25.49 & 0.41 & 0.50 & 22.75 & 0.43 & 0.50 & 37.19 & 0.33 & 0.56 & 38.71 & 0.41 & 0.49 & - & - & - \\
      Ctrl-Adapter \cite{lin2024ctrl} & SD1.5 & 8 & 20.32 & 0.62 & 0.70 & 18.51 & 0.61 & 0.75 & 21.15 & 0.54 & 0.71 & - & - & - & - & - & - & - & - & - \\
      \midrule

      % ===================== SD2.1 =====================
      ControlNet-XS \cite{zavadski2024controlnet} & SD2.1 & 50 & 16.79 & 0.65 & 0.60 & 20.11 & 0.55 & 0.52 & - & - & - & - & - & - & - & - & - & - & - & - \\
      \textbf{JDM} & SD2.1 & 1 & 20.62 & 0.46 & 0.41 & 25.06 & 0.40 & 0.38 & 24.95 & 0.28 & 0.27 & 32.84 & 0.30 & 0.27 & 30.65 & 0.33 & 0.25 & 20.73 & 0.37 & 0.34 \\
      \textbf{SDXS} & SD2.1 & 1 & 22.28 & 0.35 & 0.29 & 26.57 & 0.36 & 0.32 & 26.93 & 0.31 & 0.26 & 33.93 & 0.32 & 0.29 & 30.84 & 0.30 & 0.24 & 21.05 & 0.33 & 0.31 \\
      
      \textbf{UNet-based DeltaControl} & SD2.1 & 50 & 16.22 & 0.73 & 0.66 & 20.06 & 0.57 & 0.55 & 21.64 & 0.48 & 0.53 & 28.98 & 0.36 & 0.50 & 26.48 & 0.43 & 0.50 & 16.41 & 0.57 & 0.58 \\
      \textbf{UNet-based FlashControl} & SD2.1 & 1 & 17.82 & 0.73 & 0.63 & 22.10 & 0.58 & 0.55 & 22.81 & 0.47 & 0.52 & 29.41 & 0.32 & 0.53 & 26.63 & 0.37 & 0.52 & 17.60 & 0.59 & 0.59 \\
      \midrule

      % ===================== PixArt-alpha =====================
      RepControlNet \cite{deng2024repcontrolnetcontrolnetreparameterization} & PixArt-$\alpha$ & 50 & 16.02 & 0.81 & 0.83 & 20.50 & 0.62 & 0.71 & 22.84 & 0.55 & 0.66 & 28.62 & 0.42 & 0.60 & 26.67 & 0.48 & 0.60 & 15.50 & 0.75 & 0.84 \\
      \textbf{DiT-based DeltaControl} & PixArt-$\alpha$ & 50 & 15.19 & 0.81 & 0.84 & \textbf{18.11} & 0.64 & 0.76 & 20.42 & 0.58 & 0.68 & 26.54 & 0.46 & 0.66 & \textbf{24.50} & 0.51 & 0.66 & 14.90 & 0.73 & 0.83 \\
      \textbf{\makecell{DiT-based FlashControl \\ wo LSP loss}} & PixArt-$\alpha$ & 1 & 17.83 & 0.80 & 0.83 & 20.01 & 0.64 & 0.75 & 22.64 & 0.54 & 0.70 & 26.62 & 0.42 & 0.64 & 28.16 & 0.47 & 0.63 & 18.95 & 0.73 & 0.77 \\
      \textbf{DiT-based FlashControl} & PixArt-$\alpha$ & 1 & 16.82 & 0.82 & 0.86 & 18.29 & 0.66 & 0.77 & 20.83 & 0.58 & 0.70 & \textbf{25.62} & 0.47 & 0.66 & 26.84 & 0.52 & 0.64 & 17.68 & 0.76 & 0.79 \\
      \midrule

      % ===================== FLUX.1-schnell =====================
      UniCombine \cite{wang2025unicombine} & FLUX.1-schnell & 8 & 20.95 & 0.66 & 0.84 & 25.54 & 0.50 & 0.78 & - & - & - & - & - & - & - & - & - & 19.93 & 0.62 & 0.86 \\
      OminiControl \cite{tan2025ominicontrol} & FLUX.1-schnell & 28 & 20.14 & 0.76  & 0.83 & 21.47 & 0.61 & 0.76 & - & - & - & - & - & - & - & - & - & 15.43 & 0.76 & \textbf{0.88} \\
      OminiControl \cite{tan2025ominicontrol} & FLUX.2-Klein-4B & 28 & 20.63 & 0.73  & 0.80 & 22.02 & 0.59 & 0.73 & 23.83 & 0.53 & 0.69 & 31.63 & 0.37 & 0.59 & 27.30 & 0.46 & 0.55 & 15.92 & 0.77 & 0.83 \\
      \midrule

      % ===================== FLUX.2-Klein-4B =====================
      \textbf{DiT-based DeltaControl} & FLUX.2-Klein-4B & 28 & \textbf{12.87} & \textbf{0.92} & 0.89 & \textbf{16.57} & \textbf{0.73} & 0.77 & \textbf{16.56} & \textbf{0.70} & 0.69 & \textbf{22.23} & \textbf{0.64} & 0.58 & \textbf{20.34} & \textbf{0.68} & 0.60 & \textbf{10.16} & \textbf{0.96} & \textbf{0.91} \\
      \textbf{DiT-based FlashControl} & FLUX.2-Klein-4B & 1 & 13.64 & 0.88 & \textbf{0.89} & 16.77 & 0.70 & \textbf{0.82} & 17.83 & 0.62 & \textbf{0.76} & 23.68 & 0.49 & \textbf{0.70} & 21.21 & 0.55 & \textbf{0.72} & 12.18 & 0.95 & 0.91 \\
      \bottomrule
    \end{tabular}}
  \label{tab:updated_results}
  \vspace{-5mm}
\end{table*}

\subsection{Stage 2: One-step Controllable Generative Model FlashControl}
\label{43}

\paragraph{AutoAdapter in One-Step Distillation.}
We retain the AutoAdapter module unchanged during Stage 2, reusing it inside the distilled one-step student. 
%This architectural consistency is paramount: if the control interface used during teacher training were altered, the student would be forced to simultaneously learn one-step synthesis and a novel control representation. 
By reusing AutoAdapter, FlashControl inherits the normalized and reliability-weighted residuals optimized in Stage 1. Let the one-step student be denoted as $f_\theta(z \mid y, c)$ and the trainable fake score as $\epsilon_\psi$. In an alternating distillation framework like DMD, the gradient update follows:
$$\nabla_\theta \mathcal{L}_{\text{DMD}} = \mathbb{E}_{t, y, z, c} \left[ w(t) \left( \hat{\epsilon}_\phi - \hat{\epsilon}_\psi \right) \frac{\partial f_\theta(z \mid y, c)}{\partial \theta} \right].$$
Without AutoAdapter's alignment, control perturbations fluctuate sharply across timesteps and condition types, escalating the variance of both the student and LoRA fake score updates. By enforcing $\delta_{\text{ctrl}} = g \odot c_{\text{align}}$, the injected residual is normalized before reaching either path. Consequently, AutoAdapter actively supports faster and more stable convergence in distillation.
%by regularizing the scale and reliability of the control perturbations.

\paragraph{Latent Sensitivity Preservation (LSP).}
While score distillation provides an efficient formulation for one-step generation, we observe a distinct optimization instability specific to controllable distillation: the student rapidly becomes overly mode-seeking. It overfits to a narrow teacher trajectory, loses sensitivity to the input Gaussian noise, and collapses to a near-deterministic mapping for any given prompt-condition pair $(y, c)$. Qualitatively, this manifests as late-stage stylized collapse and oversmoothing; quantitatively, it produces sharp, disruptive spikes in both student and LoRA losses during late training. 
% To mitigate this, we introduce the \textbf{Latent Sensitivity Preservation (LSP)} objective. 

% Let $\delta z \sim \mathcal{N}(0,\sigma_{\delta}^{2} I)$ be a small latent perturbation, and define
% \[
% x_0 = f_{\theta}(z \mid y,c), \qquad x_0^{\delta} = f_{\theta}(z+\delta z \mid y,c).
% \]
% We then define
% \[
% L_{\mathrm{LSP}} =\frac{d(x_0, x_0^{\delta})}{\|\delta z\|_2 + \varepsilon},
% \]
% where $d(\cdot,\cdot)$ denotes cosine distance. The numerator measures the output change induced by a small latent perturbation, normalized by the perturbation magnitude in the input space. If the student becomes locally insensitive to its latent input, the ratio decreases and the penalty is activated.
To mitigate this, we introduce the \textbf{Latent Sensitivity Preservation (LSP)} objective. 
For two i.i.d.\ noise inputs $\mathbf{z}$ and $\mathbf{z}'$ in $\mathcal{N}(0,I)$, we extract the corresponding image latents produced by the student model when using the same text prompt $\mathbf{y}$ and condition $\mathbf{c}$:
\begin{equation}
  \mathbf{x}_0 \;=\; f_\theta(\mathbf{z} \mid \mathbf{y},\mathbf{c}), \qquad
  \mathbf{x}_0' \;=\; f_\theta(\mathbf{z}' \mid \mathbf{y},\mathbf{c}).
\end{equation}
Then, we compute the output variation $d_{\text{out}} = D(\mathbf{x}_0, \mathbf{x}_0')$ and the input variation $d_{\text{in}}  = D(\mathbf{z}, \mathbf{z}')$, given $D$ is a distance metric like $\ell_2$.
Our \textbf{LSP objective} is computed as:
\begin{equation}
  \mathcal{L}_{\text{LSP}}
  \;=\;
  -\,\frac{d_{\text{out}}}{d_{\text{in}}+\varepsilon}
\end{equation}
\noindent with $\varepsilon$ is machine epsilon to avoid division by zero. This recipe aims to maximize $\Delta$output / $\Delta$input, discourage the “flat” mapping $f_\theta(\mathbf{z})$, prevent generators from ignoring the latent noise. In one‑step distillation, the only stochastic source is $\mathbf{z}$, preserving a non‑trivial Jacobian $\frac{\partial f_{\theta}}{\partial z}$ is critical for \textbf{diversity and sharpness} \cite{mao2019modeseekinggenerativeadversarial}.

Crucially, LSP is a bounding regularization rather than an unbounded diversity maximizer. It preserves conditional faithfulness while strictly preventing degenerate one-to-one mapping collapse. The final FlashControl objective is formulated as:
$$\mathcal{L}_{\text{FlashControl}} = \mathcal{L}_{\text{DMD}} + \lambda_{\text{LSP}} \mathcal{L}_{\text{LSP}},$$
where $\mathcal{L}_{\text{DMD}}$ is the baseline distillation objective and $\lambda_{\text{LSP}}$ scales the sensitivity regularization. In summary, our complete pipeline achieves state-of-the-art one-step fidelity because AutoAdapter transforms the control signal into an easily optimizable state, while LSP guarantees the student does not undergo deterministic collapse once that control signal dominates the generative trajectory.

% \paragraph{We Drop Adversarial Distillation.} In Stage 2, we distill the single-stream model into a one-step generator by optimizing a control-preserving score distillation objective, $\mathcal{L}_{\text{DMD}}$. Crucially, we deliberately drop the adversarial (GAN) objectives standard in recent distillation frameworks like DMD22. In controllable generation, the goal is not unconstrained realism, but realism strictly subject to a spatial constraint $p(x \mid y,c)$. Adversarial signals primarily optimize the unconditional image marginal, creating a gradient conflict ($\langle \nabla_\theta \mathcal{L}_{\text{adv}}, \nabla_\theta \mathcal{L}_{\text{DMD}} \rangle < 0$) that actively pushes the student away from the control manifold. While multi-step models can correct these deviations iteratively, a one-step student possesses no corrective trajectory. By relying purely on $\mathcal{L}_{\text{DMD}}$ through our frozen AutoAdapter interface, we ensure the student receives stable, normalized control residuals ($\delta_{\text{ctrl}} = g \odot c_{\text{align}}$) without adversarial interference.

% BEGIN OLD 4.4

%% file: sec/5_exps.tex
\section{Experiments}
\label{sec:exps}
% \begin{figure}[t]
%   \centering
%   \includegraphics[width=0.0\linewidth]{sec/images/qual.pdf}
%   \caption{
%     Qualitative comparison across control modalities including: Canny, Depth, Inpainting, Hough, Pose, and Segmentation. Crossed cells mark unsupported controls.
%   }
%   \label{fig:qual_fig}
%   % \label{fig:qual}
%   \vspace{-2mm}
% \end{figure}

\begin{figure}[t]
  \centering
  \includegraphics[width=0.8\linewidth]{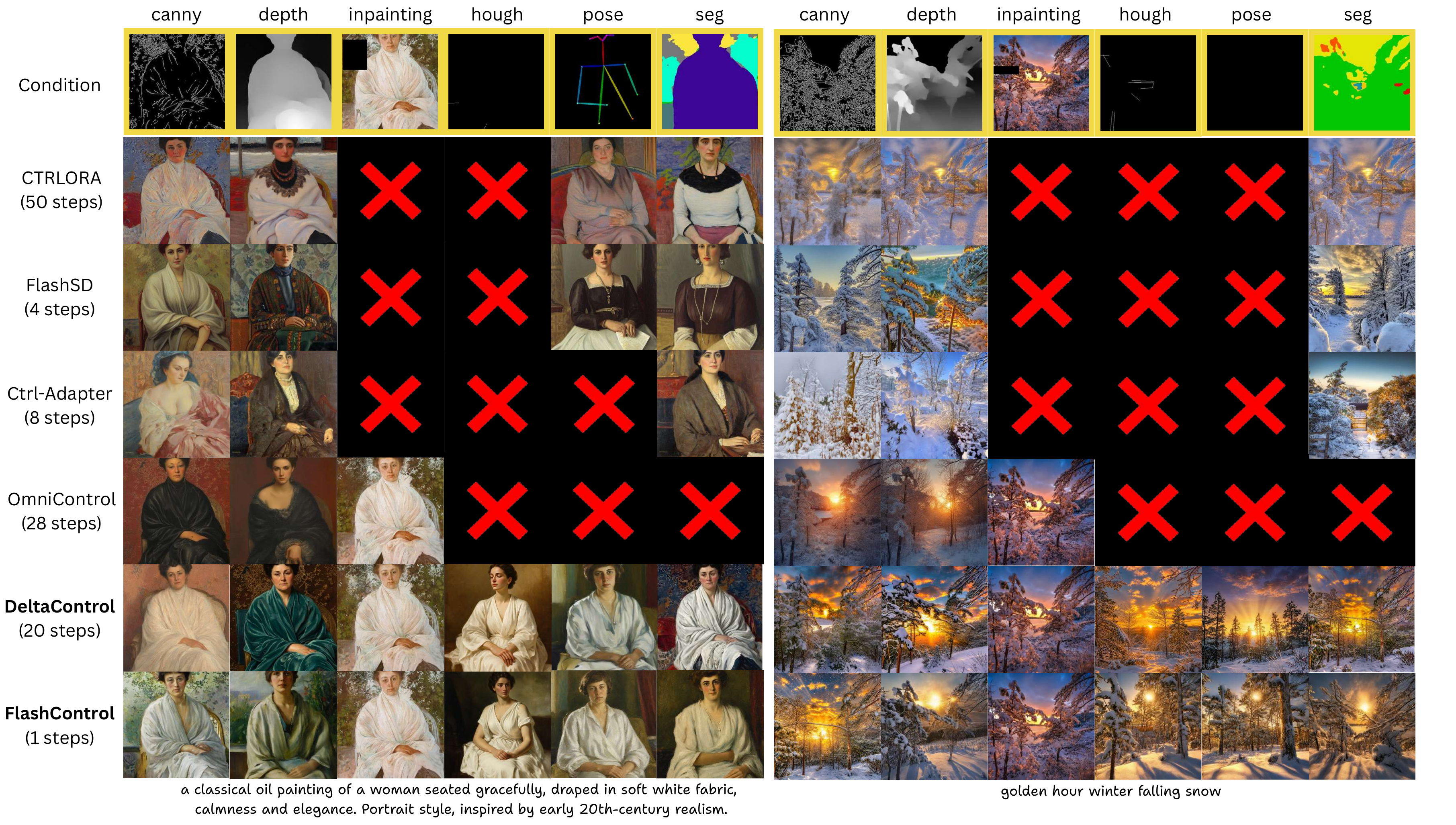}
  \caption{
    Qualitative comparison across control modalities including: Canny, Depth, Inpainting, Hough, Pose, and Segmentation. Crossed cells mark unsupported controls.
  }
  \label{fig:qual}
\end{figure}

\subsection{Experimental Setup}
\vspace{-1mm}\myheading{Training details.}
We provide complete implementation detail in Appendix \cref{abl:training_details}.

\vspace{-1mm}\myheading{Evaluation protocol.} Follow previous work, we leverage COCO2017-5k \cite{lin2015microsoftcococommonobjects} for evaluation dataset. We evaluate six control types (Canny, Depth, Segmentation, Pose, Hough lines, and Inpainting) for both multi‑step and one‑step models (1 NFE) under same setting using three metrics FID\cite{heusel2018ganstrainedtimescaleupdate}, Precision (Prec) and Recall (Rec).

\begin{table*}[!t]
  \centering
  \scriptsize
  \setlength{\tabcolsep}{2.0pt}
  \renewcommand{\arraystretch}{1.05}

  % ===================== Table 1 =====================
  \begin{minipage}[t]{0.35\textwidth}
    \centering
    \caption{\textbf{Ablation on AutoAdapter:} FID$\downarrow$ on 30k iters.}
    \label{tab:abl_autoadapter}
    \vspace{-1mm}
    \resizebox{\linewidth}{!}{%
    \begin{tabular}{l|ccc}
      \toprule
      \textbf{Method} & \textbf{Canny} & \textbf{Depth} & \textbf{Seg} \\
      \midrule
      \red{--} AutoAdapter        & 28.15 & 31.11 & 31.61 \\
      \red{--} distribution match & 26.38 & 29.51 & 30.70 \\
      \red{--} token-wise gate    & 26.88 & 29.64 & 31.19 \\
      \textbf{\green{+} AutoAdapter} & \textbf{26.03} & \textbf{29.24} & \textbf{30.62} \\
      \bottomrule
    \end{tabular}
    }
  \end{minipage}
  \hfill
  % ===================== Table 2 =====================
  \begin{minipage}[t]{0.18\textwidth}
    \centering
    \caption{\textbf{Inference time per image.}}
    \label{tab:inferspeed}
    \vspace{-1mm}
    \resizebox{\linewidth}{!}{%
    \begin{tabular}{l|r}
      \toprule
      \textbf{Method} & \textbf{s/img} \\
      \midrule
      ControlNet   & 2.1748 \\
      DeltaControl & 2.0576 \\
      FlashControl & 0.0429 \\
      \bottomrule
    \end{tabular}
    }
  \end{minipage}
  \hfill
  % ===================== Table 3 =====================
  \begin{minipage}[t]{0.43\textwidth}
    \centering
    \caption{\textbf{Training efficiency} under the same protocol with SD2.1.}
    \label{tab:train_efficiency}
    \vspace{-1mm}
    \resizebox{\linewidth}{!}{%
    \begin{tabular}{lccccc}
      \toprule
      \textbf{Method} &
      \textbf{Fwd} &
      \textbf{Bwd} &
      \textbf{Mem} &
      \textbf{Train.} &
      \textbf{Total} \\
      \midrule
      ControlNet    & 0.2689 & 0.1450 & 30GB   & 294MB & 904MB \\
      T2I-Adapter   & 0.2486 & 0.0601 & 22GB   & 83MB  & 693MB \\
      RepControlNet & 0.2710 & 0.0970 & 18.8GB & 297MB & 908MB \\
      DeltaControl  & 0.2411 & 0.0714 & 17GB   & 297MB & 610MB \\
      \bottomrule
    \end{tabular}
    }
  \end{minipage}

  \vspace{-5mm}
\end{table*}

\vspace{-1mm}\myheading{Efficiency of ControlNet vs \textbf{DeltaControl}.}
\label{tab:inferspeed}
 We quantified the training and inference efficiency of a dual‑branch ControlNet against our single‑branch \textbf{DeltaControl}. First, we set up training with the same protocol for both models, with backbone PixArt-$\alpha$ with a single condition. 
 %As reported in \cref{tab:train_efficiency}, DeltaControl reduces backward time by 50\% and training memory by 40\% compared with Controlnet. 
 We measure inference time of ControlNet, DeltaControl, and FlashControl in \cref{tab:inferspeed}. %By using only one diffusion step, FlashControl achieves impressive speed, with inference time just 0.042s in bfloat16. 

\subsection{Quantitative comparison}
\cref{tab:updated_results} compares our multi‑step teacher \textbf{DeltaControl} and one‑step student \textbf{FlashControl} against prior controllable baselines across all control types. We can observe that across tasks, our DiT‑based one‑step model matches or surpasses prior multi‑step baselines. For example, on Canny, FlashControl (FLUX.2-Klein-4B, 1 NFE) achieves \textbf{FID 13.64}, essentially closing the gap to its 28‑step teacher 12.70, while outperforming strong multi‑step baselines like OminiControl 20.14. The same scenario happens for other conditionings. This indicates that the distillation preserves fidelity and coverage. We also surpass other controllable one-step baselines like JDM \cite{luo2025addingadditionalcontrolonestep} and SDXS \cite{song2024sdxsrealtimeonesteplatent} too.

% \vspace{-4mm}
\subsection{Qualitative comparison}
We compare \textbf{DeltaControl} and \textbf{FlashControl} (PixArt-$\alpha$) with recent controllable generators: CTRLORA, FlashSD, Ctrl-Adapter and OminiControl. We evaluate six common conditioning types. For each method we use its recommended sampling process.
In \cref{fig:qual}, for the portrait, FlashControl keeps consistent brushwork and fabric detail. For outdoor scenes, it maintains coherent lighting and snow particle density, avoiding over-sharpening typical of aggressive control. DeltaControl and FlashControl produce aligned image with high perceptual quality, whereas several baselines introduce color shifts around the hole boundary. Despite using 1 step, FlashControl delivers quality on par with multi-step methods (e.g., 28–50 steps), demonstrating controllability in real time. Additionally, we also provide control fidelity benchmarking in Appendix.

\subsection{Ablation study}
\vspace{-1mm}\myheading{AutoAdapter ablation.}
We study AutoAdapter on 3 conditions (Canny, Depth, Seg), comparing:
(i) no adapter, (ii) w/o distribution matching, (iii) w/o token-wise gate, and
(iv) full AutoAdapter (Tab~\ref{tab:abl_autoadapter}).
Full AutoAdapter gives the lowest FID and fastest convergence for all conditions.
Removing either component degrades FID,
and removing the adapter entirely yields the worst and less stable training. We also study how AutoAdapter make distillation converge better in \cref{abl:aa_benefit}.

\vspace{-1mm}\myheading{LSP in distillation.}
We ablate the LSP loss in FlashControl to test diversity/sharpness.
Without it, FID worsens (Tab.~\ref{tab:updated_results}), suggesting the loss helps avoid
near-deterministic collapse and keeps the one-step model sensitive to input noise.
We also compare dual-branch ControlNet vs.\ our single-branch \textbf{DeltaControl} under the same
PixArt-$\alpha$ setup (single condition): DeltaControl reduces backward time by 50\% and training
memory by 40\% (Tab.~\ref{tab:train_efficiency}).
Finally, FlashControl is fastest at inference, achieving 0.042s per $512\times512$ image on an
NVIDIA A100 40GB (bfloat16) (Tab \cref{tab:abl_autoadapter}). Additionally, we examine how LSP fight collapse in distillation in \cref{abl:lsp_fight}.

\vspace{-1mm}\myheading{Varying Teacher Architecture.}
Tab.~\ref{tab:teacher_arch_distillation} isolates a single factor: the teacher architecture. All rows use the same one-step distillation pipeline with varying teacher model. This comparison shows that the improvement does not come from the distillation recipe alone. Branch-based teachers such as ControlNet and T2I-Adapter transfer poorly, confirming discussions in \cref{42}. More importantly, even among \emph{single-stream} teachers, \textbf{DeltaControl consistently yields the strongest one-step student}. This shows that single-stream design by itself is insufficient: the teacher must be structured specifically for distillation. RepControlNet improves inference-time structure through reparameterization, and OminiControl provides a minimal universal controller for DiT backbones, but DeltaControl is the one that exposes the cleanest teacher-to-student distillation interface.

\begin{table}[t]
\centering
\small
\caption{
Effect of teacher architecture under the same one-step distillation pipeline.
All rows use the same distillation recipe; only the teacher architecture changes.
Metric is FID on COCO-5k.
}
\label{tab:teacher_arch_distillation}
\vspace{2pt}
\setlength{\tabcolsep}{4.5pt}
\renewcommand{\arraystretch}{1.12}
\resizebox{0.7\linewidth}{!}{%
\begin{tabular}{@{}l l cccccc@{}}
\toprule
\rowcolor{headergray}
\textbf{Teacher Architecture} &
\textbf{Backbone} &
\textbf{Canny} $\downarrow$ &
\textbf{Depth} $\downarrow$ &
\textbf{Seg} $\downarrow$ &
\textbf{Pose} $\downarrow$ &
\textbf{Hough} $\downarrow$ &
\textbf{Inpainting} $\downarrow$ \\
\midrule

ControlNet    & SD2.1 & 47.15 & 49.12 & 52.62 & 58.03 & 53.01 & 31.74 \\
T2I-Adapter   & SD2.1 & 24.72 & 29.62 & 28.51 & 37.80 & 33.72 & 24.15 \\
RepControlNet & SD2.1 & 21.41 & 25.29 & 26.40 & 34.33 & 31.01 & 21.23 \\
\rowcolor{pastelgreen}
DeltaControl (Ours) 
              & SD2.1 & \textbf{19.53} & \textbf{24.27} & \textbf{24.93} & \textbf{32.62} & \textbf{29.13} & \textbf{19.07} \\
\midrule

OminiControl  & FLUX.2-Klein-4B & 17.38 & 23.01 & 25.42 & 32.51 & 30.55 & 17.40 \\
\rowcolor{pastelgreen}
DeltaControl (Ours) 
              & FLUX.2-Klein-4B & \textbf{11.27} & \textbf{16.95} & \textbf{24.63} & \textbf{30.91} & \textbf{29.45} & \textbf{15.94} \\
\bottomrule
\end{tabular}%
}
\vspace{-5mm}
\end{table}

\subsection{Extended Alignment Metrics for Quantitative Results}
In addition to the primary alignment metrics presented in the main paper including FID, Precision, and Recall, we also include task‑specific metrics to more clearly evaluate how well the model adheres to different conditioning inputs. Follow \cite{sun2024anycontrol}, we evaluate structural consistency using RMSE for both canny edge and depth conditions, and measure semantic alignment for segmentation conditions using mean Pixel Accuracy (mPA).
\begin{table}[ht]
    \centering
    \caption{Comparison of model performance on COCO-5K using extended alignment metrics. ``-'' denotes that condition is unavailable.}
    \small
    \begin{tabular}{llp{2.5cm}p{2.2cm}p{2.2cm}}
        \hline
        \textbf{Models} & Backbone & \textbf{Canny (RMSE $\downarrow$)} & \textbf{Depth (RMSE $\downarrow$)} & \textbf{Seg (mPA $\uparrow$)} \\
        \hline
        ControlNet & SD1.5 & 87.30 & 36.49 & 0.12 \\
        CTRLORA & SD1.5 & 82.52 & 33.59 & 0.16 \\
        CtrlAdapter & SD1.5 & 84.05 & 38.43 & 0.13 \\
        DeltaControl & SD2.1 & 76.25 & 34.75 & 0.16 \\
        FlashControl & SD2.1 & 76.64 & 34.12 & 0.18 \\
        \midrule
        RepControlNet & PixArt-$\alpha$ & 87.56 & 32.77 & 0.19 \\
        DeltaControl & PixArt-$\alpha$ & 76.63 & 31.30 & 0.19 \\
        FlashControl & PixArt-$\alpha$ & 75.32 & \textbf{28.63} & \textbf{0.20} \\
        \midrule
        UniCombine & FLUX.1-schnell & 75.35 & 37.30 & - \\
        OminiControl & FLUX.2-Klein-4B & \textbf{74.83} & 32.74 & 0.16 \\
        DeltaControl & FLUX.2-Klein-4B & 75.78 & 30.72 & 0.22 \\
        FlashControl & FLUX.2-Klein-4B & 76.84 & \textbf{29.24} & \textbf{0.23} \\
        \hline
    \end{tabular}
    \label{tab:model_performance}
\end{table}

%% file: sec/6_conclusion.tex
\vspace{-3mm}\section{Conclusion}
FlashControl shows that single-step, controllable image generation is possible with a streamlined, single-branch design. It preserves much of the teacher’s controllability while simplifying both training and inference. In evaluations, it performs on par with traditional two-branch methods. For future work, we can investigate simultaneous multi-condition composition (e.g., combining depth and pose in a single input).
% Further improvements include supporting a wider range of control types and broadening the evaluation scope.
% Future work should investigate principled scaling strategies, adapter placement, and improved normalization.

%% file: sec/X_suppl.tex
\clearpage
\newpage

\begin{figure}[t]
  \centering
  \includegraphics[width=0.7\linewidth]{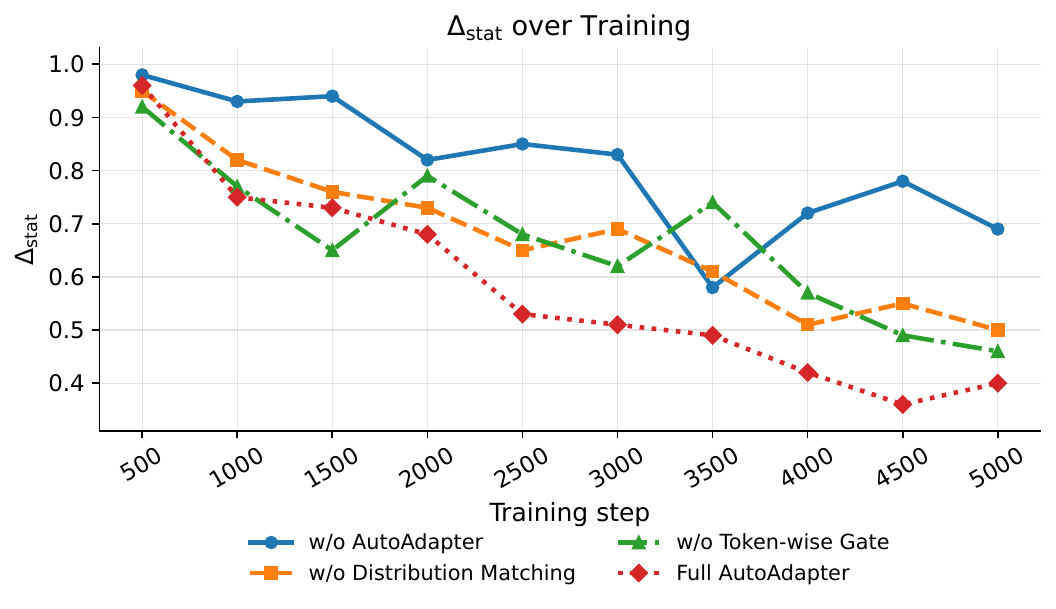}
   \caption{Evolution of the control-hidden statistic mismatch \(\Delta_{\mathrm{stat}}\) during
training. All variants start from a similar level, but the full AutoAdapter reduces
the mismatch fastest and maintains the lowest \(\Delta_{\mathrm{stat}}\) throughout training.
Removing AutoAdapter leads to the highest mismatch, while removing either
distribution matching or the token-wise gate gives intermediate performance.}

  \label{fig:delta_stat}
\end{figure}

\begin{figure}
    \centering
    \begin{subfigure}[t]{0.48\columnwidth}
        \centering
        \includegraphics[width=\linewidth]{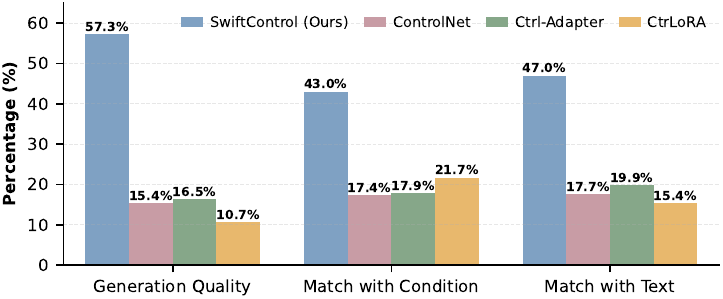}
        % \caption{}
        \label{fig:maze}
    \end{subfigure}\hfill
    \begin{subfigure}[t]{0.48\columnwidth}
        \centering
        \includegraphics[width=\linewidth]{sec/images/user_study.pdf}
        % \caption{}
        \label{fig:user_study}
    \end{subfigure}
    \caption{User study. Our FlashControl achieves higher perceptual quality and $50\times$ faster.}
\end{figure}

\paragraph{Appendix Overview.}
This appendix is organized as follows. We first provide extended quantitative alignment metrics for Canny, depth, and segmentation conditions in \cref{tab:model_performance}. We then analyze the training behavior of one-step distillation, including the benefits of AutoAdapter in \cref{abl:aa_benefit}, the stabilizing effect of LSP in \cref{abl:lsp_fight}, and the feature-alignment analysis in \cref{fig:delta_stat}. Next, we present the user study in \cref{fig:user_study}, followed by related work in \cref{sec:related}, dataset construction, and training implementation details in \cref{abl:training_details}. We further study the effect of the number of training conditions in \cref{fig:num_cond} and compare different distillation objectives in \cref{tab:tab6}. Finally, we provide additional experimental details in \cref{params_choosing}, discuss limitations in \cref{sec:limitations}, describe AutoAdapter in detail in \cref{app:auto_adapter}, and show more qualitative results in \cref{fig:supp_qual1,fig:supp_qual2,fig:supp_qual3,fig:supp_qual4,fig:supp_qual5,fig:supp_qual6}.

\begin{figure}[th]
  \centering

  % ---------- Left: Fig A ----------
  \begin{minipage}[t]{0.36\linewidth}
    \centering
    \includegraphics[width=\linewidth]{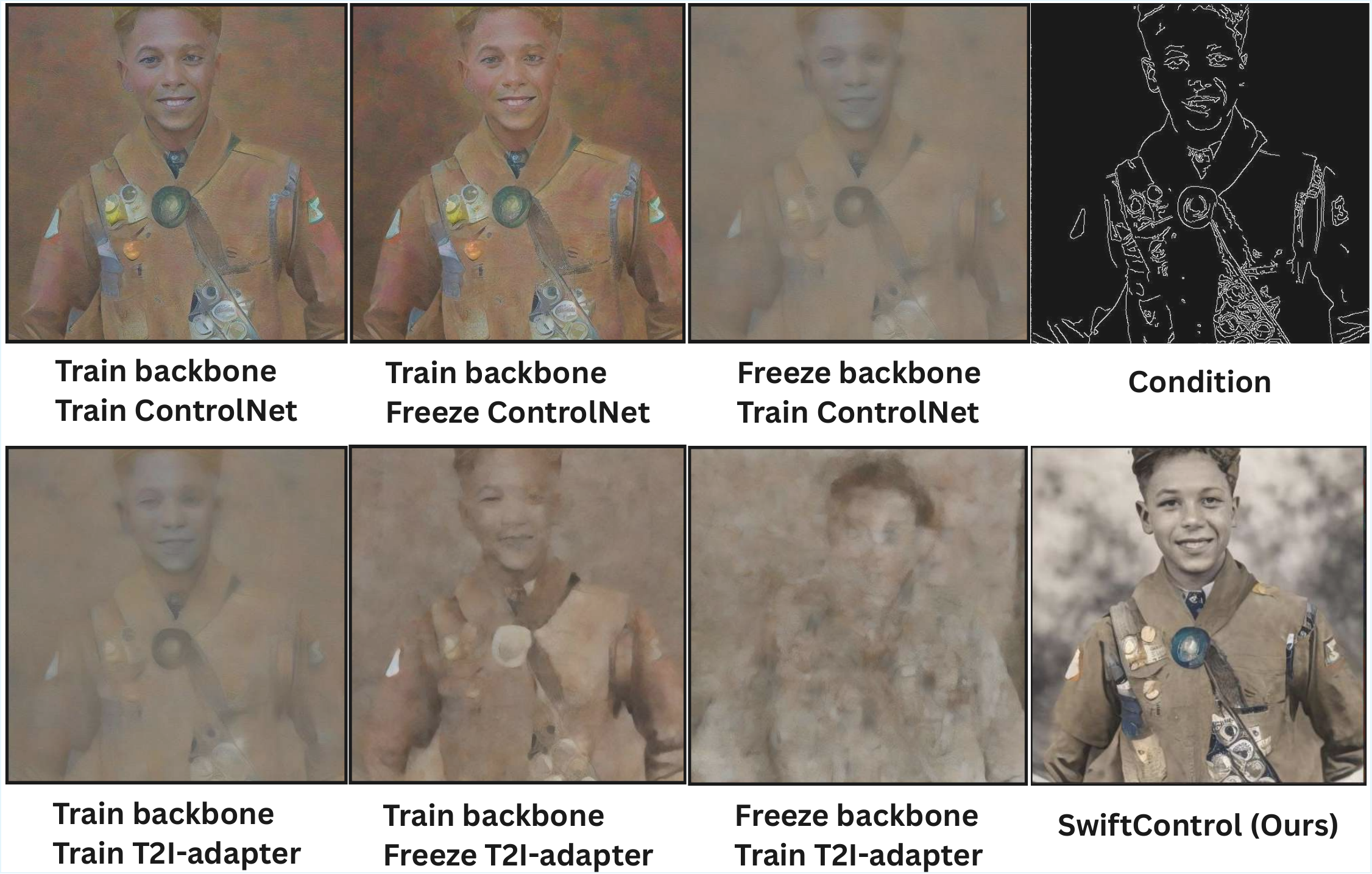}
    \captionof{figure}{One‑step distillation with ControlNet, T2I-Adapter vs DeltaControl.}
    \label{fig:controlnet_fail}
  \end{minipage}\hfill
  % ---------- Right: Fig X (contains B and C) ----------
  \begin{minipage}[t]{0.63\linewidth}
    \centering
    \begin{subfigure}[t]{0.49\linewidth}
      \centering
      \includegraphics[width=\linewidth]{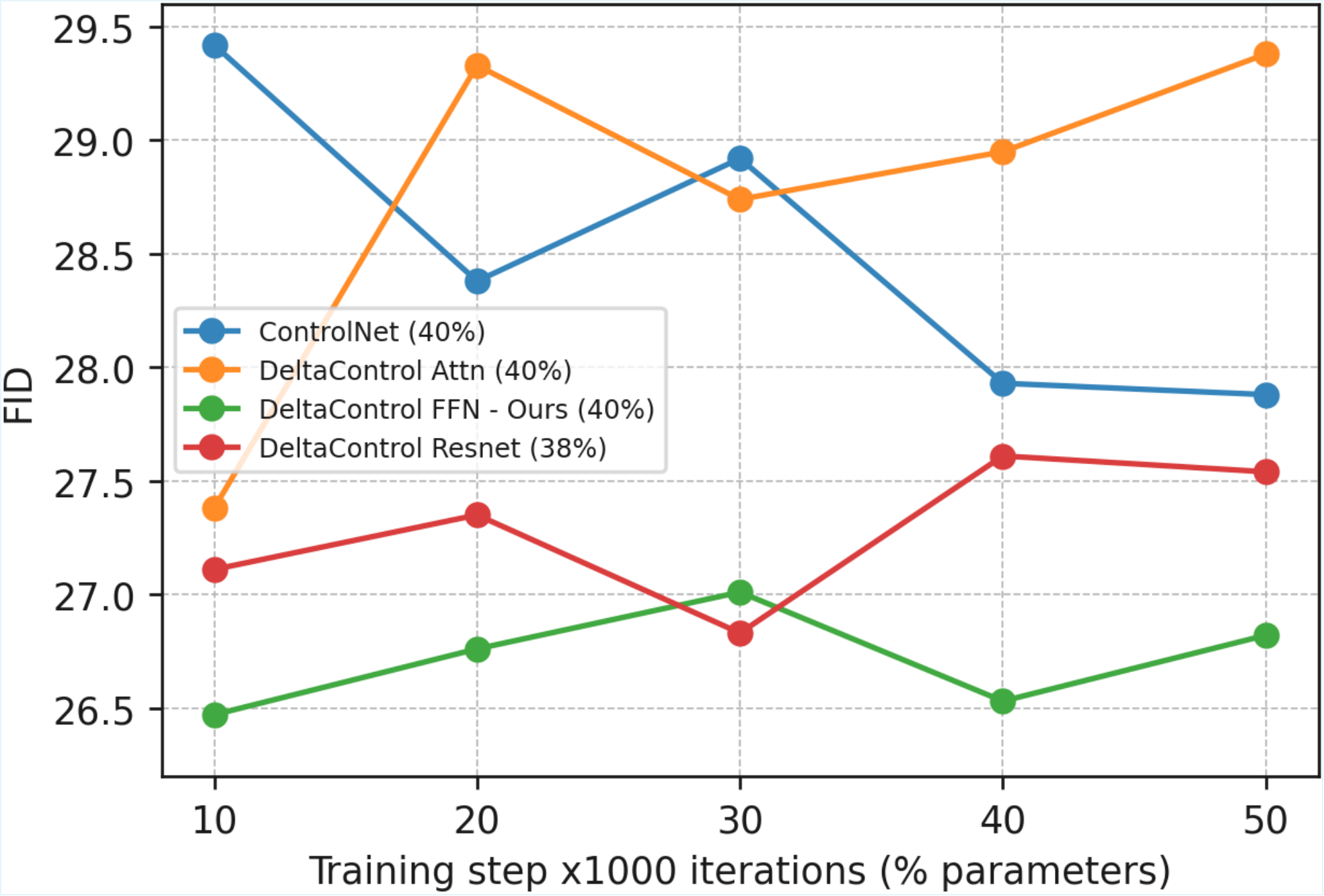}
      \caption{UNet (SD2.1)}
      \label{fig:unet_params}
    \end{subfigure}\hfill
    \begin{subfigure}[t]{0.49\linewidth}
      \centering
      \includegraphics[width=\linewidth]{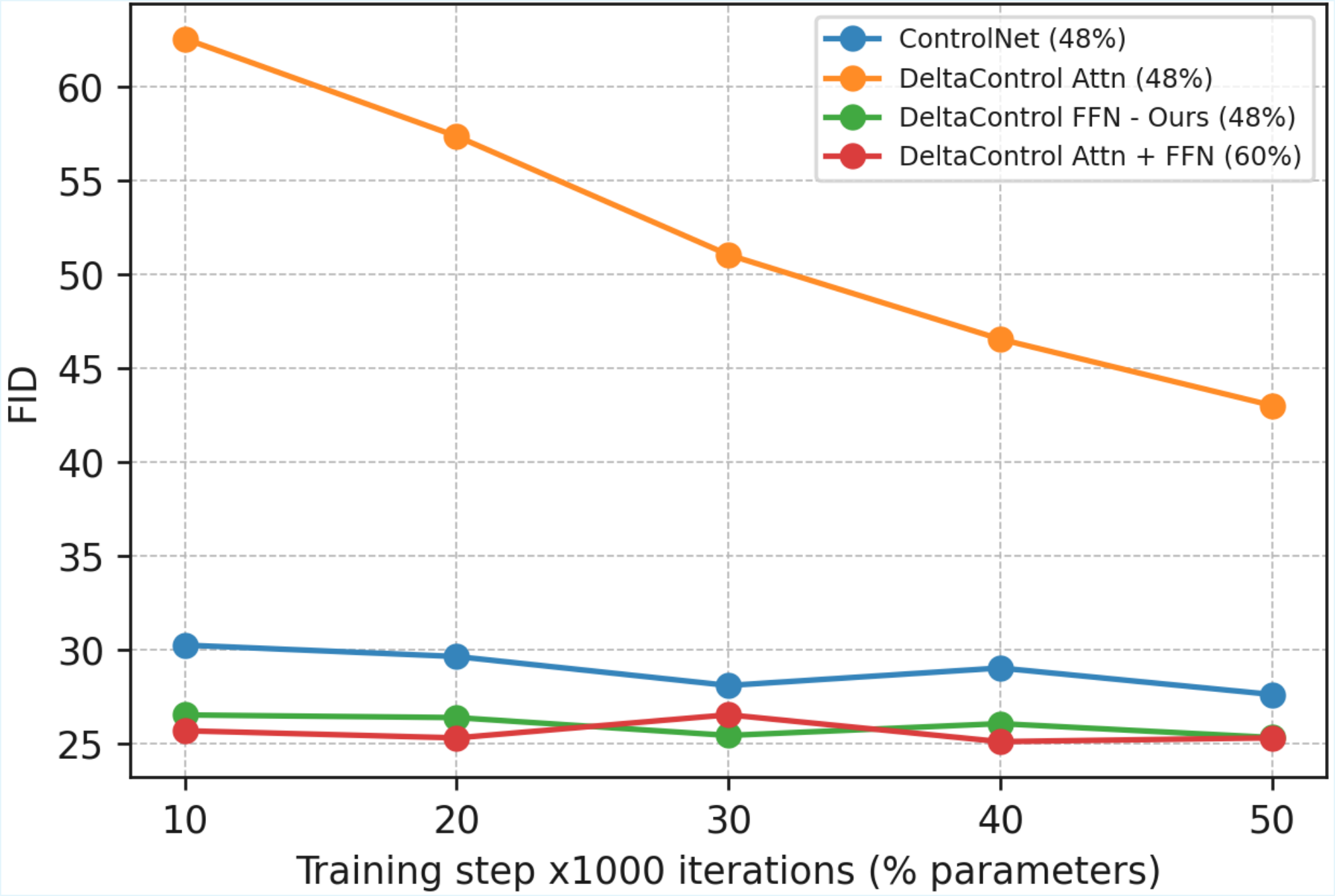}
      \caption{DiT (PixArt-$\alpha$)}
      \label{fig:dit_params}
    \end{subfigure}

    \captionof{figure}{Effect of selective fine‑tuning on convergence (lower is better)}
    \label{fig:vary_params}
  \end{minipage}
  % \vspace{-8mm}
\end{figure}

\begin{figure}[t]
  \centering
  \includegraphics[width=\linewidth]{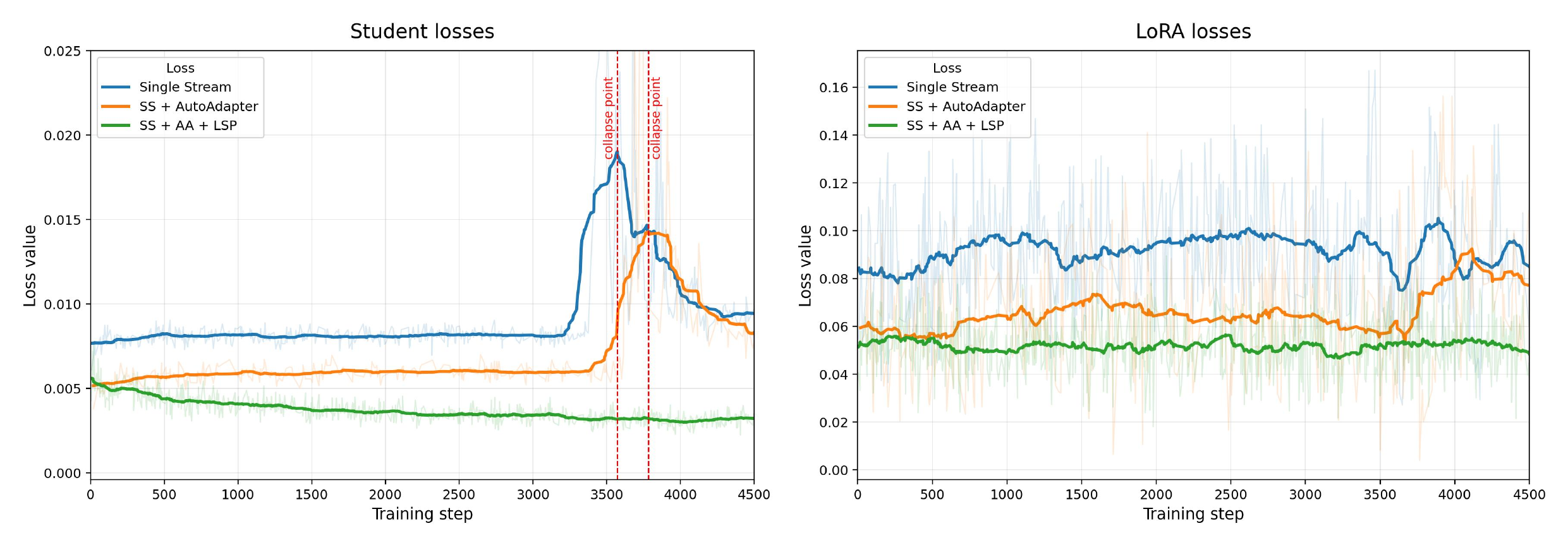}
\caption{Training losses during one-step distillation. We compare \textit{Single stream}, \textit{SS (Single Stream) + AutoAdapter}, and \textit{SS + AA (AutoAdapter) + LSP} using both student loss and LoRA-teacher loss across training steps. AutoAdapter makes optimization smoother and lowers the losses compared with the plain single-stream baseline, while adding LSP further suppresses the loss spike around 3.5k--4k steps, indicating more stable distillation.}
  \label{fig:evi_figure}
\end{figure}

\begin{figure}[t]
  \centering
  \includegraphics[width=0.7\linewidth]{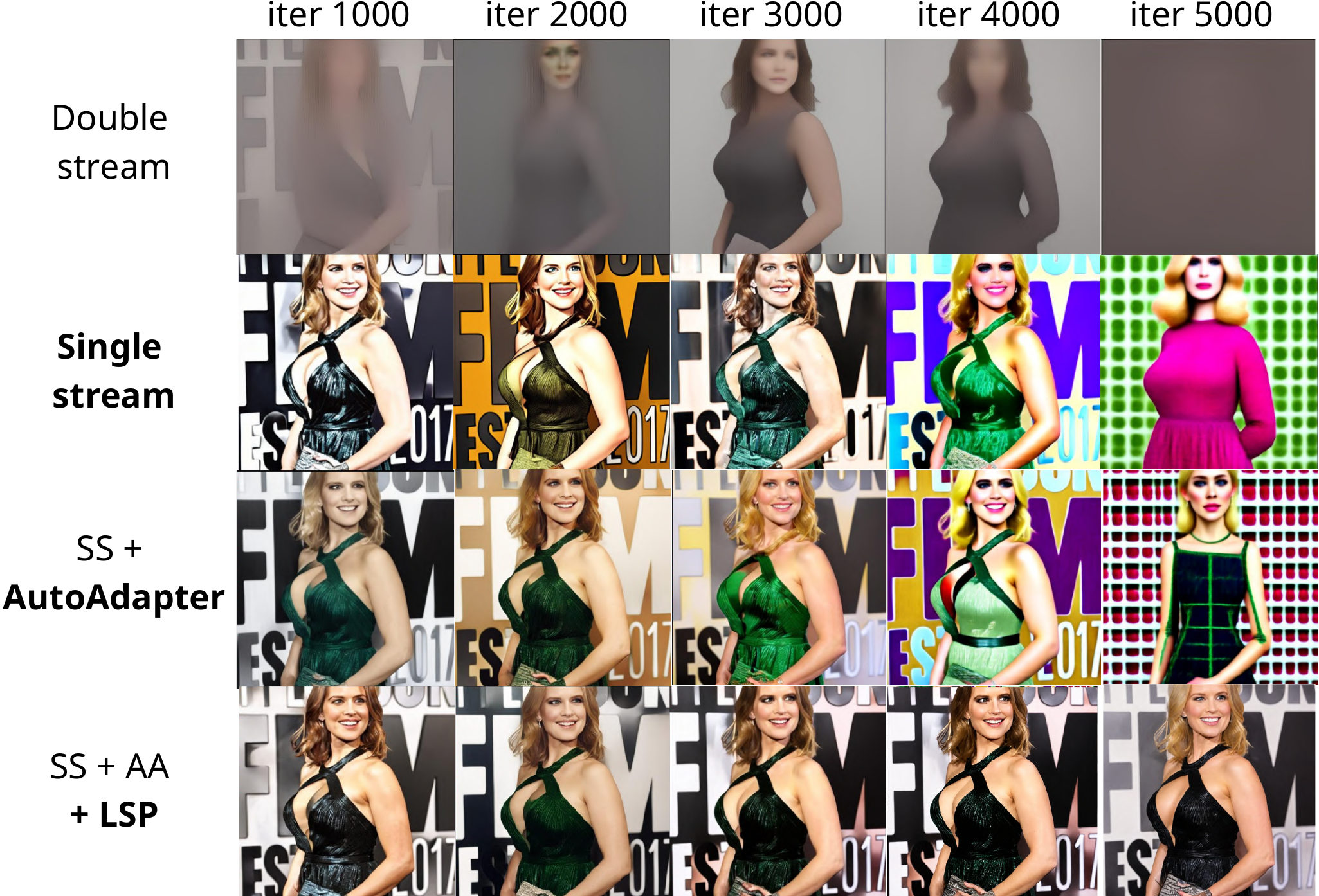}
   \caption{Qualitative training at different distillation iterations. AutoAdapter improves visual convergence over the plain single-stream baseline, and adding LSP further stabilizes late-stage training and avoids collapse. SS is Single-Stream and AA is AutoAdapter.}

  \label{fig:evi_visual}
\end{figure}

\section{AutoAdapter benefits distillation.}
\label{abl:aa_benefit}
We investigate whether AutoAdapter actively benefits the one-step distillation process, rather than merely strengthening the Stage-1 teacher. We compare the optimization trajectories of the plain single-stream baseline, \texttt{SS + AutoAdapter}, and \texttt{SS + AA + LSP} under identical distillation conditions. Visually, AutoAdapter establishes recognizable structure much earlier in training and yields consistently cleaner conditional outputs \cref{fig:evi_visual}. Quantitatively, it produces lower and substantially smoother loss curves prior to the late-stage instability region \cref{fig:evi_figure}. These findings confirm that AutoAdapter transcends simple Stage-1 quality improvements: by providing a statistically well-conditioned control interface, it makes the alternating Stage-2 updates inherently easier and more stable to optimize.

\section{LSP benefits distillation.}
\label{abl:lsp_fight}
We examine the training dynamics to isolate the stabilizing effect of LSP. As shown in \ref{fig:evi_figure}, distillation without LSP exhibits a pronounced spike in both student and LoRA losses. This quantitative instability manifests visually in \ref{fig:evi_figure}, where the ablated model drifts into severe degradation and oversaturationin training. Conversely, the inclusion of LSP yields flat, stable loss curves and preserves coherent visual structure throughout the same training range. These findings substantiate our core theoretical claim: LSP functions strictly as a collapse-prevention mechanism. Rather than acting as a generic diversity heuristic, it actively preserves the student's sensitivity to input noise, successfully preventing the model from degenerating into a near-deterministic mapping once the spatial control signal dominates.

\section{Effect of AutoAdapter on feature alignment.}
We further analyze AutoAdapter by tracking the statistic mismatch
\(\Delta_{\mathrm{stat}}\) during training. As shown in Fig.~\ref{fig:delta_stat},
all variants begin from a similar starting point, but the full AutoAdapter reduces
the mismatch more quickly and remains consistently lowest throughout training.
Removing AutoAdapter gives the highest \(\Delta_{\mathrm{stat}}\) over nearly the whole
trajectory, showing that naive condition injection leaves a persistent mismatch
between the control features and the backbone hidden states. Removing either
distribution matching or the token-wise gate leads to intermediate behavior:
both variants improve over no adapter, but both remain worse than the full model.
Overall, this result supports our claim that AutoAdapter improves controllable
generation by making the injected control features more compatible with the
backbone feature space. All experiments are employed with the same backbone SD2.1, loss DMD only and hyperparameters to ensure fairness.

\section{User Study}
    We conducted a user study with 279 participants to evaluate perceptual quality. Each participant compared outputs from four methods (FlashControl, ControlNet, Ctrl-Adapter, CtrLoRA) under the same condition and prompt, selecting the best image based on Image Quality, Condition Alignment, and Textual Alignment. We used 30 prompts (10 each for Canny, Depth, and Segmentation). As shown in \cref{fig:user_study}, FlashControl was preferred in \textbf{57.3\%} of cases for Image Quality, \textbf{43.0\%} for Condition Alignment, and \textbf{47.0\%} for Textual Alignment, outperforming all baselines despite using only \textbf{1 NFE} versus 50 for others. This shows that our one-step method achieves superior perceptual quality while being $50\times$ faster.

\myheading{Instant Controllable Diffusion Models.}
Several recent works share the same goal as ours: one-step controllable diffusion generation. SDXS~\cite{song2024sdxsrealtimeonesteplatent} distills a controllable student from a multi-step teacher, but still shows a noticeable quality and diversity gap compared to strong multi-step controllable models. JDM~\cite{luo2025addingadditionalcontrolonestep} provides a principled joint distribution matching objective over images and conditions, but its training pipeline is more complex, has been mainly examined with a single UNet backbone, and does not directly address the teacher-architecture mismatch that arises when compressing controllable generation into one step. Additionally, neither method has a publicly available training implementation.

\section{Dataset.}
To train a controllable DeltaControl, we use a 2M subset of the Re-LAION5B dataset \cite{schuhmann2022laionb} as supervised images for consistency with other baseline works. Following ControlNet \cite{zhang2023adding}, we extract corresponding conditions from the dataset, including Canny Edge \cite{canny1986computational}, Depth Map \cite{ranftl2020towards}, Segmentation Mask \cite{zhou2017scene}, M-LSD lines \cite{gu2021realtime}, Openpose \cite{cao2019openpose} and Inpainting. We extract all control modalities from the same LAION images by following the ControlNet preprocessing pipeline on their official GitHub. Some prior works use different condition extraction pipelines (e.g., similar canny settings but different depth estimators), so we follow the ControlNet to unify condition generation across all modalities. For FlashControl distillation, we use only caption and conditions since no image is required. 

\section{Training Implementation Details}
\label{abl:training_details}
We summarize the implementation details for Stage 1 \textbf{DeltaControl} and Stage 2 \textbf{FlashControl} in \cref{tab:hparams} across three backbones Stable Diffusion 2.1, PixArt-$\alpha$ and FLUX.2-klein-4B. For training LoRA layers, we apply LoRA to all FFN modules described in \cref{params_choosing} for Stable Diffusion 2.1 and PixArt-$\alpha$. For FLUX.2-klein-4B, we follow the same principle: LoRA is applied to all FFN layers in the double-stream blocks and to the first 14 fused projection layers in the single-stream blocks, so that the total number of trainable parameters remains comparable to the ControlNet setup.
\begin{table*}[t]
\label{abl:}
    \centering
    \scriptsize
    \caption{Training hyperparameters and compute settings for Stage 1 \textbf{DeltaControl} and Stage 2 \textbf{FlashControl}. For the conditioning projectors, we randomly initialize these weights and also train them using LoRA.}
    \label{tab:training_settings}
    \label{tab:hparams}
    \begin{tabular}{l c c c}
        \toprule
        \multicolumn{4}{c}{Stage 1 \textbf{DeltaControl}} \\
        \midrule
        Backbone & Stable Diffusion 2.1 & PixArt-$\alpha$ & FLUX.2-klein-4B \\
        \# of training data & \multicolumn{3}{c}{2M supervised images} \\
        Condition types & \multicolumn{3}{c}{Canny, Depth, Segmentation, Hough, Pose, Inpainting} \\
        \# of LoRA parameters &  412M & 370M & 190M \\
        \# of AutoAdapter parameters & 4M & 4M & 30M \\
        \# Total parameter & 614M & 864M & 3.9B \\
        LoRA rank & 512 & 512 & 64 \\
        Learning rate & \multicolumn{3}{c}{$\epsilon=5 * 10^{-5}$} \\
        Optimizer & \multicolumn{3}{c}{Adam ($\beta_1=0$, $\beta_2=0.999$, $\epsilon=10^{-8}$)} \\
        Resolution & \multicolumn{3}{c}{512} \\
        \midrule
        \# of GPUs & \multicolumn{3}{c}{8$\times$H100 (80G)} \\
        Batch size per GPU & \multicolumn{3}{c}{16} \\
        Training Time (Hours) & \multicolumn{3}{c}{3} \\
        Seed & \multicolumn{3}{c}{42} \\

        \midrule
        \multicolumn{4}{c}{Stage 2 \textbf{FlashControl}} \\
        \midrule
        Backbone & Stable Diffusion 2.1 & PixArt-$\alpha$ & FLUX.2-klein-4B \\
        \midrule
        Paradigm
        & Diffusion
        & Diffusion
        & \makecell[c]{Rectified Flow} \\
        Fake score LoRA rank & 64 & 64 & 128 \\
        Student LoRA rank & 256 & 256 & 128 \\
        Resolution & \multicolumn{3}{c}{512} \\
        Teacher Guidance CFG & \multicolumn{3}{c}{3} \\
        Optimizer & \multicolumn{3}{c}{Adam ($\beta_1=0$, $\beta_2=0.999$, $\epsilon=10^{-8}$)} \\
        % $\lambda_{\mathrm{CLIP}}$ & \multicolumn{3}{c}{0.1} \\
        % Target CLIP score & \multicolumn{3}{c}{0.4} \\
        $\lambda_{\mathrm{LSP}}$ & \multicolumn{3}{c}{0.1} \\
        Diversity Distance & \multicolumn{3}{c}{Cosine} \\
        \midrule
        \# of GPUs & \multicolumn{3}{c}{8$\times$H100 (80G)} \\
        Batch size per GPU & \multicolumn{3}{c}{8} \\
        Training Time (Hours) & \multicolumn{3}{c}{4} \\
        Seed & \multicolumn{3}{c}{42} \\
        \bottomrule
    \end{tabular}
\end{table*}

\begin{figure}[t]
    \centering
    \includegraphics[width=0.6\linewidth]{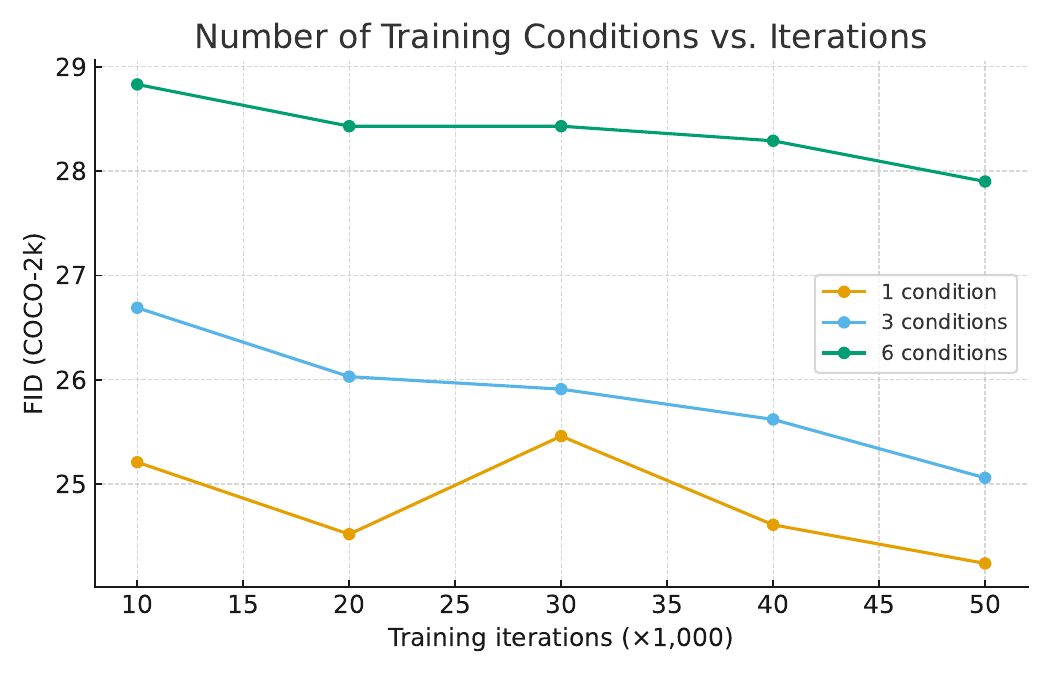}
    \caption{\textbf{Training Difficulty}. With more control types and under the same budget, FID and convergence rate are slightly worse.}
    \label{fig:num_cond}
\end{figure}

\section{Number of Training Conditions}

To see whether training with more control types degrades general quality, we vary the number of condition types seen during training (1, 3, and 6) and track FID during training with the same evaluation protocol, as in \cref{fig:num_cond}. The result is that, with the same model size and training budget, more conditions hurt image quality. This may be because giving the model more condition types forces its parameters to explain a broader set of inputs.

\begin{table}[t]
\caption{Quantitative results (FID) on different distillation methods across 6 tasks. FlashControl leverages DMD.}
\centering
\resizebox{0.7\linewidth}{!}{%
\begin{tabular}{lcccccc}
\toprule
        & Canny & Depth & Seg & Pose & Hough & Inpainting \\
\midrule
DMD PixArt-$\alpha$ (\textbf{Ours}) & 17.83 & 20.10 & 22.64 & 26.62 & 28.16 & 18.95 \\
DMD2                  & 16.45 & 19.29 & 21.29 & 26.38 & 28.95 & 19.02 \\
\bottomrule
\end{tabular}
}
\label{tab:tab6}
\end{table}

\section{Varying Distillation Methods}

To study how the choice of distillation method affects our framework, we replace DMD with DMD2. Both configs have LSD loss. With adversarial loss, DMD2 bring more robust performance compared with DMD.

\section{More Details About Our Experiments}
\label{params_choosing}

\paragraph{Selective parameter fine-tuning experiment.}
In \cref{fig:vary_params}, we compare ControlNet with our DeltaControl variants on both a U-Net--based model (Stable Diffusion 2.1) and a DiT-based model (PixArt-$\alpha$).

PixArt-$\alpha$ has 28 transformer blocks; for each DeltaControl variant:
\begin{itemize}
    \item \textbf{Attn}: fine-tune all parameters in the self-attention and cross-attention layers of all 28 blocks.
    \item \textbf{FFN}: fine-tune all feedforward layers in all 28 blocks.
    \item \textbf{Attn+FFN}: fine-tune a subset of parameters taken from both the Attn and FFN variants, again across all 28 blocks.
\end{itemize}

Stable Diffusion 2.1 has 4 down blocks, 1 mid block, and 4 up blocks; for each DeltaControl variant:
\begin{itemize}
    \item \textbf{ResNet}: fine-tune all convolution layers in the ResNet sub-blocks of all down blocks, the mid block, and the first two up blocks.
    \item \textbf{FFN}: fine-tune all convolution layers in the ResNet sub-blocks of all down blocks and the mid block, plus all feedforward layers in the transformer blocks.
    \item \textbf{Attn}: fine-tune all convolution layers in the ResNet sub-blocks of all down blocks, the mid block, and the first up block, plus all self-attention and cross-attention layers in the transformer blocks.
\end{itemize}

We choose the number of tuned layers so that the percentage of trainable parameters matches ControlNet for a fair comparison. All models are trained with a single Canny edge condition: we (i) encode the condition with a VAE, (ii) map it with a small trainable embedding module, and (iii) add it to the intermediate latent at the network head in the same way for all variants.

\section{Limitations}
\label{sec:limitations}

FlashControl has several limitations. First, although the final student is fast at inference, the method still requires a strong pretrained backbone and a two-stage teacher--student training pipeline, which introduces non-trivial training cost. Second, our experiments focus on $512 \times 512$ images and six common spatial controls, so we do not claim direct generalization to higher-resolution synthesis, video, or specialized domains. Third, FlashControl currently handles one control type at a time. Our results suggest that training with more condition types under the same budget can slow convergence and slightly degrade quality, so robust multi-condition composition remains future work. Finally, like other controllable image generators, fast one-step inference may lower the cost of creating misleading visual content; responsible release should include usage restrictions, provenance tools, and safety filters.

\section{More Qualitative Results}

\begin{figure}[ht]
    \centering
    \includegraphics[width=\linewidth]{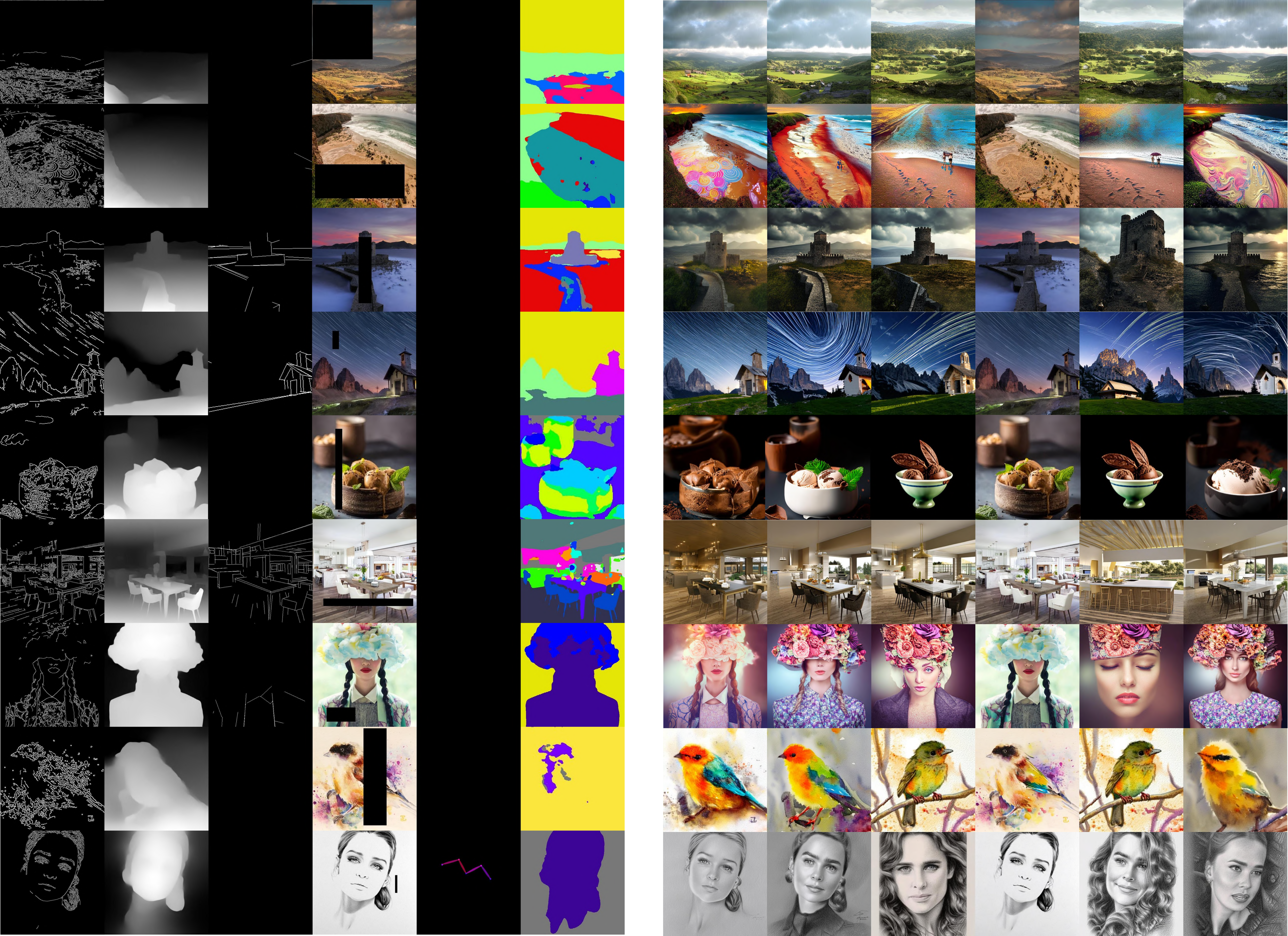}
    \caption{Qualitative results of DeltaControl PixArt-$\alpha$ (20 steps). Zoom in for a better view.}
    \label{fig:supp_qual1}
\end{figure}

\begin{figure}[ht]
    \centering
    \includegraphics[width=\linewidth]{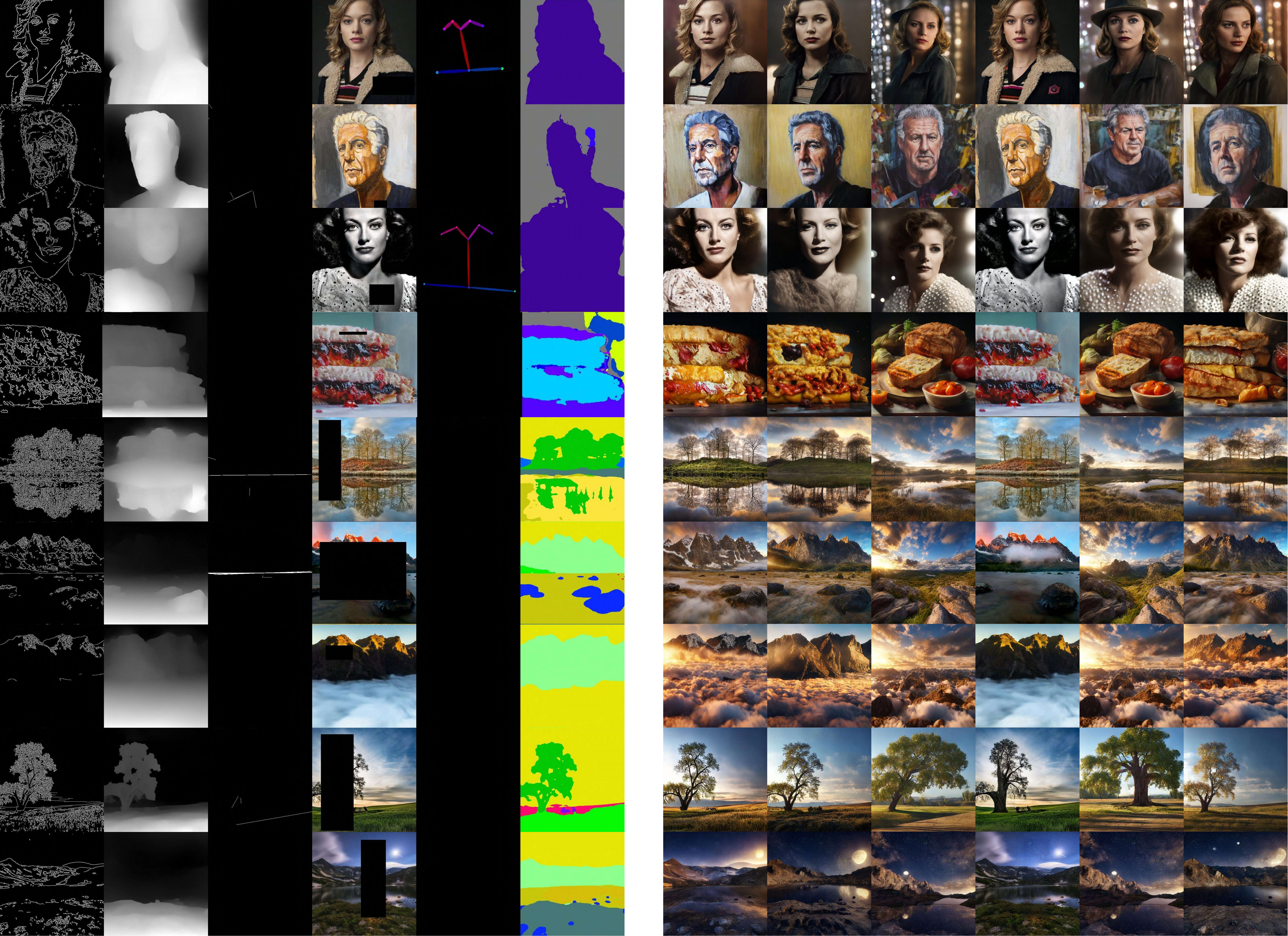}
    \caption{Qualitative results of FlashControl PixArt-$\alpha$ (1 step). Zoom in for a better view.}
    \label{fig:supp_qual2}
\end{figure}

\begin{figure}[ht]
    \centering
    \includegraphics[width=\linewidth]{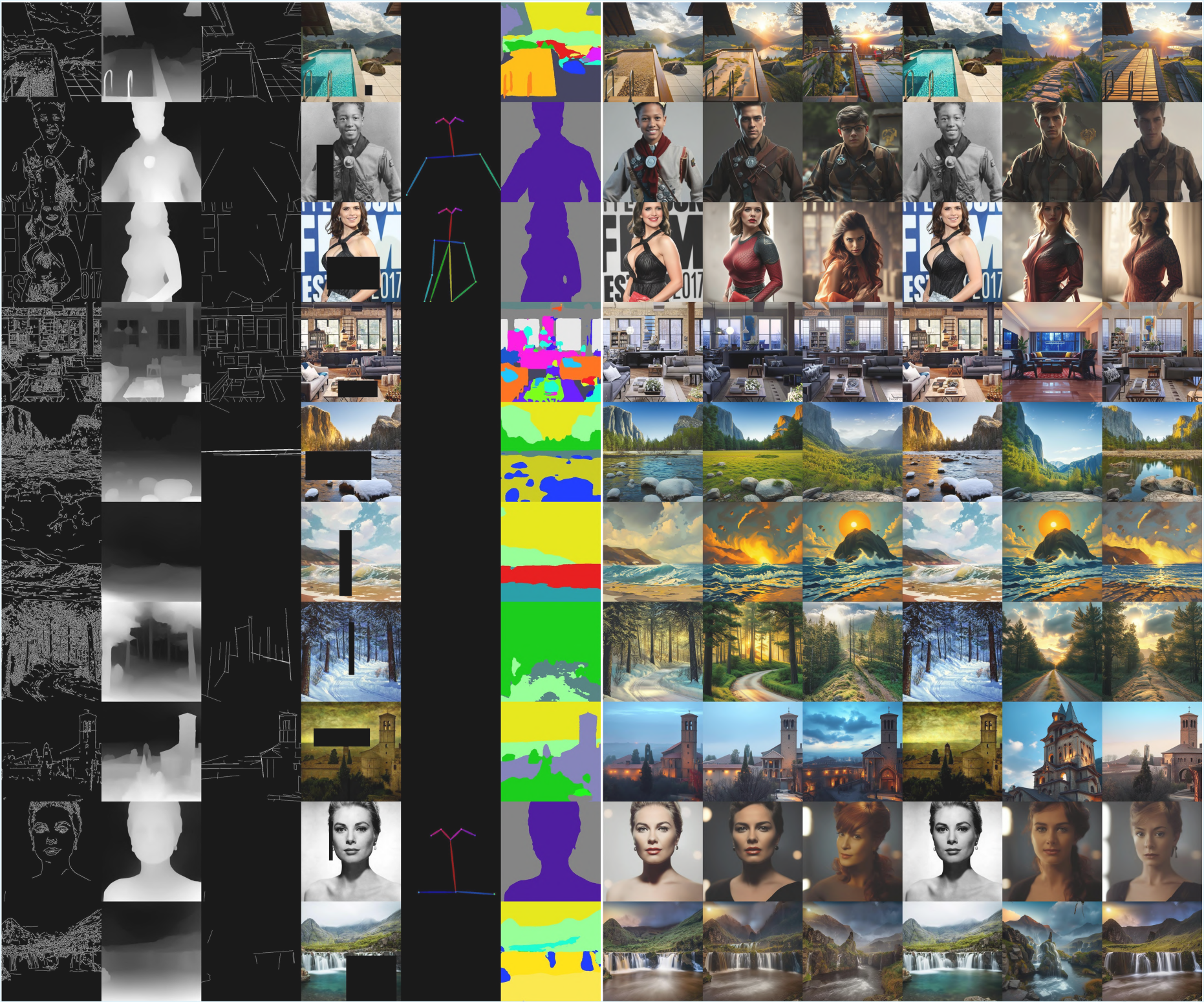}
    \caption{Qualitative results of DeltaControl FLUX.2-klein-4B (20 steps). Zoom in for a better view.}
    \label{fig:supp_qual3}
\end{figure}

\begin{figure}[ht]
    \centering
    \includegraphics[width=\linewidth]{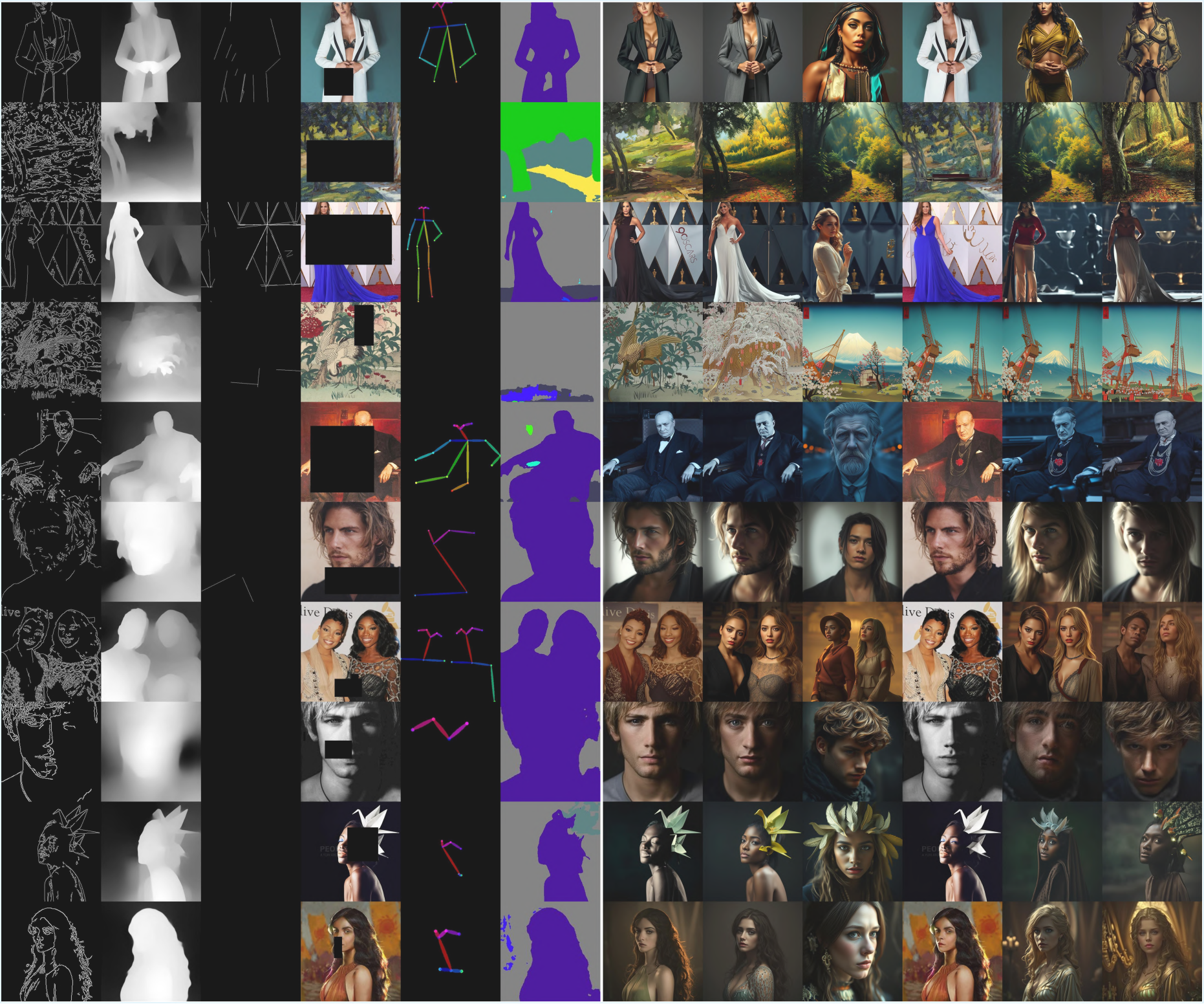}
    \caption{Qualitative results of DeltaControl FLUX.2-klein-4B (20 steps). Zoom in for a better view.}
    \label{fig:supp_qual4}
\end{figure}

\begin{figure}[ht]
    \centering
    \includegraphics[width=\linewidth]{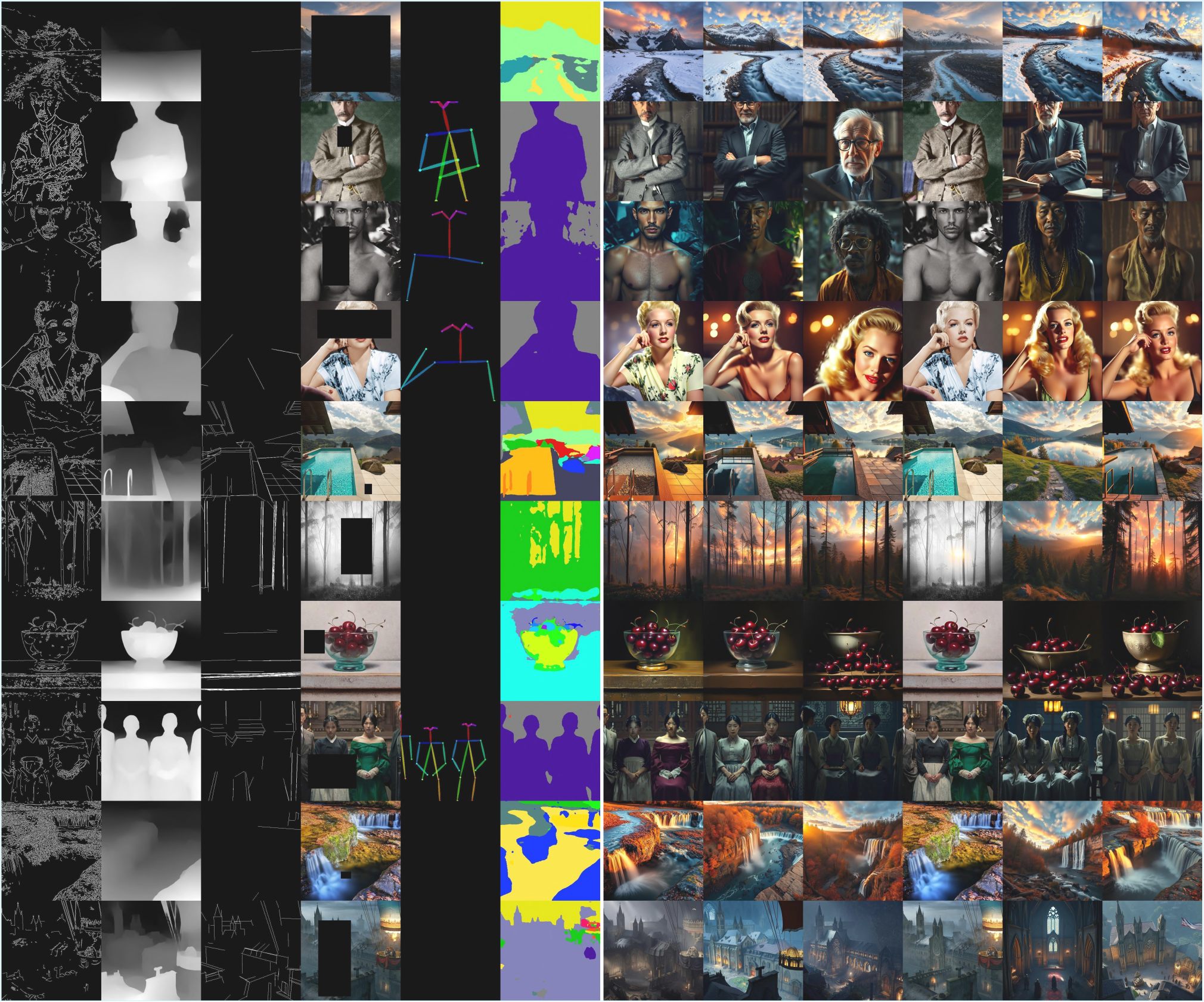}
    \caption{Qualitative results of FlashControl FLUX.2-klein-4B (1 step). Zoom in for a better view.}
    \label{fig:supp_qual5}
\end{figure}

\begin{figure}[ht]
    \centering
    \includegraphics[width=\linewidth]{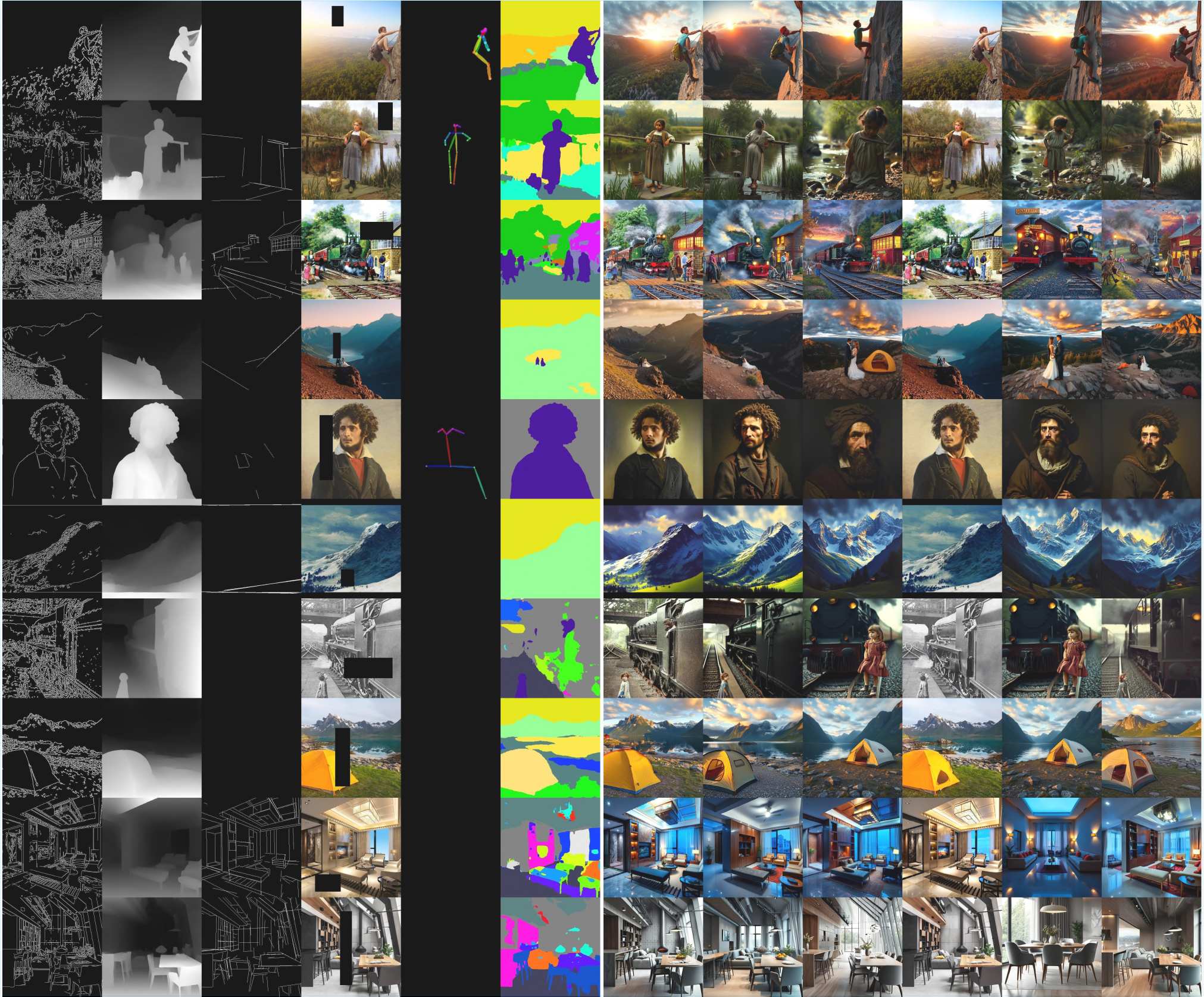}
    \caption{Qualitative results of FlashControl FLUX.2-klein-4B (1 step). Zoom in for a better view.}
    \label{fig:supp_qual6}
\end{figure}

More qualitative results for DeltaControl and FlashControl are shown in \cref{fig:supp_qual1,fig:supp_qual2,fig:supp_qual3,fig:supp_qual4,fig:supp_qual5,fig:supp_qual6}.

\section{AutoAdapter: Detailed Design}
\label{app:auto_adapter}

In this section, we provide a more detailed explanation of the \emph{AutoAdapter} module introduced in Section~4.2 of the main paper. Recall that AutoAdapter is a lightweight head that converts a spatial control input (\eg, a Canny edge map or a depth map) into token features that can be safely injected into the single-flow backbone of DeltaControl.

\paragraph{Overall data flow.}
Given a single control image $c_{\mathrm{in}}$ and backbone hidden states $h \in \mathbb{R}^{B \times N \times D}$ at the network head, AutoAdapter proceeds in three high-level stages:
\begin{enumerate}
    \item Encode the condition image into a latent embedding $c_0$ using the same VAE encoder as the base diffusion model, then project it to the same dimension as $h$.
    \item Match the distribution of the condition tokens to the hidden-state tokens using the \emph{distribution matching module} (whitening and re-scaling, as defined in Eqs.~(5)--(7) of the main paper).
    \item Modulate each condition token with a \emph{token-wise uncertainty gate}, and add the gated condition features to the backbone hidden states.
\end{enumerate}

This design lets the model interpret control inputs in the same feature space as the backbone, while automatically deciding \emph{where} and \emph{how strongly} each control token should affect the current latent. We provide pseudo-code for the AutoAdapter module in \cref{alg:auto_adapter}.

\paragraph{Distribution matching in practice.}
We (i) VAE-encode $c_{\mathrm{in}}$ to $c_0$, (ii) whiten per-channel over tokens as in Eq.~(5), (iii) feed the whitened tokens, the current hidden-state tokens, and the timestep embedding $\phi(t)$ into a small MLP to predict per-channel scale and shift parameters $(s, b)$ as in Eq.~(6), and (iv) re-scale and shift to produce $c_{\text{align}}$ as in Eq.~(7).

Intuitively, this module performs an AdaIN-style alignment between the condition stream and the backbone stream. Without this step, directly adding raw condition tokens to $h$ can either drown out the original features (if the condition is too large) or make the control nearly invisible (if it is too small). The distribution matching module keeps the condition in a ``safe'' range where gradients remain stable across timesteps and conditions.

\paragraph{Token-wise uncertainty gating.}
Even after distribution matching, not every condition token should affect the hidden state equally. For example, a poorly extracted edge map or an occluded region should be trusted less than a clean, well-aligned region. AutoAdapter therefore applies a token-wise gate $g_i$ to each condition token $c_{\text{align}, i}$ before adding it to the corresponding hidden token $h_i$.

The gate $g_i$ is computed from three signals:
\begin{enumerate}
    \item A similarity term between the hidden token and the aligned condition token (high similarity $\Rightarrow$ gate opens).
    \item A time-dependent factor $\text{time}(t)$ that encourages stronger control at early denoising steps and weaker control near the final steps.
    \item An uncertainty term produced by a small MLP on $c_{\text{align}, i}$, which decreases the gate when the condition token appears noisy or unreliable.
\end{enumerate}
These three terms are combined into a single scalar gate and squashed by a sigmoid, as described in Sec.~4.2.2. The gated condition token is then added to the hidden state:
\[
  h_i \leftarrow h_i + g_i \cdot c_{\text{align}, i}.
\]
In regions where the condition aligns well with the current latent (\eg, edges or poses that agree with the generated structure), $g_i$ is close to $1$ and the control strongly influences the hidden state. When the condition is inconsistent or noisy, $g_i$ shrinks toward $0$, effectively turning the control off in those tokens.

\begin{algorithm}[!ht]
    \caption{AutoAdapter flow}
    \label{alg:auto_adapter}
    \begin{algorithmic}
        \REQUIRE Hidden states $h \in \mathbb{R}^{B \times N \times D}$ at the backbone head,
                 condition image $c_{\text{in}}$ (single active condition),
                 timestep $t$.
        \ENSURE Updated hidden states $\tilde{h}$ with injected control.
        \STATE \textbf{// Encode condition as latent tokens}
        \STATE $c_0 \gets \text{Enc}(c_{\text{in}})$ \hfill\COMMENT{Same VAE encoder as the base model}
        \STATE $c_0 \gets \text{Flatten}(c_0)$ \hfill\COMMENT{Shape: $B \times \tilde{N} \times D$}
        \STATE \textbf{// Distribution matching }
        \STATE $c \gets \text{WhitenCondition}(c_0)$
               \hfill\COMMENT{Per-channel whitening as in Eq.~(5)}
        \STATE $\phi_t \gets \text{TimestepEmbedding}(t)$
        \STATE $(s, b) \gets \text{MLP}_{\text{align}}(h, c, \phi_t)$
               \hfill\COMMENT{Per-channel scales and shifts as in Eq.~(6)}
        \STATE $c_{\text{align}} \gets \text{RescaleCondition}(c, s, b)$
               \hfill\COMMENT{Aligned condition tokens as in Eq.~(7)}
        \STATE \textbf{// Token-wise uncertainty gating}
        \FOR{$i = 1$ \TO $N$}
            \STATE $h_i \gets \text{Norm}(h_i)$
                   \hfill\COMMENT{Optional: e.g., $\ell_2$-normalization}
            \STATE $c_i \gets c_{\text{align}, i}$ \hfill\COMMENT{Token aligned to hidden index $i$}
            \STATE $\text{sim}_i \gets \text{CosineSimilarity}(h_i, c_i)$
            \STATE $\tau_i \gets \text{TimeFactor}(t)$
                   \hfill\COMMENT{Stronger at early timesteps, weaker at late timesteps}
            \STATE $u_i \gets \text{MLP}_{\text{uncertainty}}(c_i)$
                   \hfill\COMMENT{Predicts token-wise uncertainty}
            \STATE $g_i \gets \text{GateFromSignals}(\text{sim}_i, \tau_i, u_i)$
                   \hfill\COMMENT{Combine signals and apply sigmoid as in Sec.~4.2.2}
            \STATE $h_i \gets h_i + g_i \cdot c_i$
        \ENDFOR
        \STATE $\tilde{h} \gets h$
        \STATE \textbf{return} $\tilde{h}$
    \end{algorithmic}
\end{algorithm}

\clearpage

%% file: checklist.tex
\section*{NeurIPS Paper Checklist}

\begin{enumerate}

\item {\bf Claims}
    \item[] Question: Do the main claims made in the abstract and introduction accurately reflect the paper's contributions and scope?
    \item[] Answer: \answerYes{}
    \item[] Justification: The abstract and introduction accurately state that the paper studies one-step controllable generation using a distillation-friendly single-stream teacher. The claimed contributions are supported by the method, quantitative comparisons, ablations, and efficiency analysis in the main paper and appendix.
    \item[] Guidelines:
    \begin{itemize}
        \item The answer \answerNA{} means that the abstract and introduction do not include the claims made in the paper.
        \item The abstract and/or introduction should clearly state the claims made, including the contributions made in the paper and important assumptions and limitations. A \answerNo{} or \answerNA{} answer to this question will not be perceived well by the reviewers. 
        \item The claims made should match theoretical and experimental results, and reflect how much the results can be expected to generalize to other settings. 
        \item It is fine to include aspirational goals as motivation as long as it is clear that these goals are not attained by the paper. 
    \end{itemize}

\item {\bf Limitations}
    \item[] Question: Does the paper discuss the limitations of the work performed by the authors?
    \item[] Answer: \answerYes{}
    \item[] Justification: The paper includes a limitations section discussing the two-stage training cost, reliance on strong pretrained backbones, evaluation at $512 \times 512$, single-control inference, multi-condition scaling, and potential misuse of fast controllable generation.
    \item[] Guidelines:
    \begin{itemize}
        \item The answer \answerNA{} means that the paper has no limitation while the answer \answerNo{} means that the paper has limitations, but those are not discussed in the paper. 
        \item The authors are encouraged to create a separate ``Limitations'' section in their paper.
        \item The paper should point out any strong assumptions and how robust the results are to violations of these assumptions (e.g., independence assumptions, noiseless settings, model well-specification, asymptotic approximations only holding locally). The authors should reflect on how these assumptions might be violated in practice and what the implications would be.
        \item The authors should reflect on the scope of the claims made, e.g., if the approach was only tested on a few datasets or with a few runs. In general, empirical results often depend on implicit assumptions, which should be articulated.
        \item The authors should reflect on the factors that influence the performance of the approach. For example, a facial recognition algorithm may perform poorly when image resolution is low or images are taken in low lighting. Or a speech-to-text system might not be used reliably to provide closed captions for online lectures because it fails to handle technical jargon.
        \item The authors should discuss the computational efficiency of the proposed algorithms and how they scale with dataset size.
        \item If applicable, the authors should discuss possible limitations of their approach to address problems of privacy and fairness.
        \item While the authors might fear that complete honesty about limitations might be used by reviewers as grounds for rejection, a worse outcome might be that reviewers discover limitations that aren't acknowledged in the paper. The authors should use their best judgment and recognize that individual actions in favor of transparency play an important role in developing norms that preserve the integrity of the community. Reviewers will be specifically instructed to not penalize honesty concerning limitations.
    \end{itemize}

\item {\bf Theory assumptions and proofs}
    \item[] Question: For each theoretical result, does the paper provide the full set of assumptions and a complete (and correct) proof?
    \item[] Answer: \answerNA{}
    \item[] Justification: The paper does not present formal theoretical results or proofs. The mathematical formulations and first-order arguments are used to motivate the method, while the main claims are empirical.
    \item[] Guidelines:
    \begin{itemize}
        \item The answer \answerNA{} means that the paper does not include theoretical results. 
        \item All the theorems, formulas, and proofs in the paper should be numbered and cross-referenced.
        \item All assumptions should be clearly stated or referenced in the statement of any theorems.
        \item The proofs can either appear in the main paper or the supplemental material, but if they appear in the supplemental material, the authors are encouraged to provide a short proof sketch to provide intuition. 
        \item Inversely, any informal proof provided in the core of the paper should be complemented by formal proofs provided in appendix or supplemental material.
        \item Theorems and Lemmas that the proof relies upon should be properly referenced. 
    \end{itemize}

    \item {\bf Experimental result reproducibility}
    \item[] Question: Does the paper fully disclose all the information needed to reproduce the main experimental results of the paper to the extent that it affects the main claims and/or conclusions of the paper (regardless of whether the code and data are provided or not)?
    \item[] Answer: \answerYes{}
    \item[] Justification: The paper describes the architecture, training stages, objectives, backbones, control modalities, datasets, evaluation protocol, and implementation details needed to reproduce the main experiments.
    \item[] Guidelines:
    \begin{itemize}
        \item The answer \answerNA{} means that the paper does not include experiments.
        \item If the paper includes experiments, a \answerNo{} answer to this question will not be perceived well by the reviewers: Making the paper reproducible is important, regardless of whether the code and data are provided or not.
        \item If the contribution is a dataset and\slash or model, the authors should describe the steps taken to make their results reproducible or verifiable. 
        \item Depending on the contribution, reproducibility can be accomplished in various ways. For example, if the contribution is a novel architecture, describing the architecture fully might suffice, or if the contribution is a specific model and empirical evaluation, it may be necessary to either make it possible for others to replicate the model with the same dataset, or provide access to the model. In general. releasing code and data is often one good way to accomplish this, but reproducibility can also be provided via detailed instructions for how to replicate the results, access to a hosted model (e.g., in the case of a large language model), releasing of a model checkpoint, or other means that are appropriate to the research performed.
        \item While NeurIPS does not require releasing code, the conference does require all submissions to provide some reasonable avenue for reproducibility, which may depend on the nature of the contribution. For example
        \begin{enumerate}
            \item If the contribution is primarily a new algorithm, the paper should make it clear how to reproduce that algorithm.
            \item If the contribution is primarily a new model architecture, the paper should describe the architecture clearly and fully.
            \item If the contribution is a new model (e.g., a large language model), then there should either be a way to access this model for reproducing the results or a way to reproduce the model (e.g., with an open-source dataset or instructions for how to construct the dataset).
            \item We recognize that reproducibility may be tricky in some cases, in which case authors are welcome to describe the particular way they provide for reproducibility. In the case of closed-source models, it may be that access to the model is limited in some way (e.g., to registered users), but it should be possible for other researchers to have some path to reproducing or verifying the results.
        \end{enumerate}
    \end{itemize}

\item {\bf Open access to data and code}
    \item[] Question: Does the paper provide open access to the data and code, with sufficient instructions to faithfully reproduce the main experimental results, as described in supplemental material?
    \item[] Answer: \answerNo{}
    \item[] Justification: At submission time, we do not provide anonymized code and checkpoints. We plan to release code, configurations, and checkpoints upon publication, subject to the policies of our company.
    \item[] Guidelines:
    \begin{itemize}
        \item The answer \answerNA{} means that paper does not include experiments requiring code.
        \item Please see the NeurIPS code and data submission guidelines (\url{https://neurips.cc/public/guides/CodeSubmissionPolicy}) for more details.
        \item While we encourage the release of code and data, we understand that this might not be possible, so \answerNo{} is an acceptable answer. Papers cannot be rejected simply for not including code, unless this is central to the contribution (e.g., for a new open-source benchmark).
        \item The instructions should contain the exact command and environment needed to run to reproduce the results. See the NeurIPS code and data submission guidelines (\url{https://neurips.cc/public/guides/CodeSubmissionPolicy}) for more details.
        \item The authors should provide instructions on data access and preparation, including how to access the raw data, preprocessed data, intermediate data, and generated data, etc.
        \item The authors should provide scripts to reproduce all experimental results for the new proposed method and baselines. If only a subset of experiments are reproducible, they should state which ones are omitted from the script and why.
        \item At submission time, to preserve anonymity, the authors should release anonymized versions (if applicable).
        \item Providing as much information as possible in supplemental material (appended to the paper) is recommended, but including URLs to data and code is permitted.
    \end{itemize}

\item {\bf Experimental setting/details}
    \item[] Question: Does the paper specify all the training and test details (e.g., data splits, hyperparameters, how they were chosen, type of optimizer) necessary to understand the results?
    \item[] Answer: \answerYes{}
    \item[] Justification: The paper reports the training data, backbones, control types, LoRA ranks, learning rates, optimizer, resolution, batch size, GPU type, training time, seed, and evaluation protocol in the main paper and appendix.
    \item[] Guidelines:
    \begin{itemize}
        \item The answer \answerNA{} means that the paper does not include experiments.
        \item The experimental setting should be presented in the core of the paper to a level of detail that is necessary to appreciate the results and make sense of them.
        \item The full details can be provided either with the code, in appendix, or as supplemental material.
    \end{itemize}

\item {\bf Experiment statistical significance}
    \item[] Question: Does the paper report error bars suitably and correctly defined or other appropriate information about the statistical significance of the experiments?
    \item[] Answer: \answerNo{}
    \item[] Justification: We do not report error bars or multi-seed confidence intervals because training multiple large diffusion teachers and one-step students is computationally expensive. Instead, we use fixed evaluation sets and matched training protocols for controlled comparisons.
    \item[] Guidelines:
    \begin{itemize}
        \item The answer \answerNA{} means that the paper does not include experiments.
        \item The authors should answer \answerYes{} if the results are accompanied by error bars, confidence intervals, or statistical significance tests, at least for the experiments that support the main claims of the paper.
        \item The factors of variability that the error bars are capturing should be clearly stated (for example, train/test split, initialization, random drawing of some parameter, or overall run with given experimental conditions).
        \item The method for calculating the error bars should be explained (closed form formula, call to a library function, bootstrap, etc.)
        \item The assumptions made should be given (e.g., Normally distributed errors).
        \item It should be clear whether the error bar is the standard deviation or the standard error of the mean.
        \item It is OK to report 1-sigma error bars, but one should state it. The authors should preferably report a 2-sigma error bar than state that they have a 96\% CI, if the hypothesis of Normality of errors is not verified.
        \item For asymmetric distributions, the authors should be careful not to show in tables or figures symmetric error bars that would yield results that are out of range (e.g., negative error rates).
        \item If error bars are reported in tables or plots, the authors should explain in the text how they were calculated and reference the corresponding figures or tables in the text.
    \end{itemize}

\item {\bf Experiments compute resources}
    \item[] Question: For each experiment, does the paper provide sufficient information on the computer resources (type of compute workers, memory, time of execution) needed to reproduce the experiments?
    \item[] Answer: \answerYes{}
    \item[] Justification: The paper reports the GPU type, GPU memory, batch size, resolution, and training time for the main Stage-1 and Stage-2 experiments, as well as training and inference efficiency measurements.
    \item[] Guidelines:
    \begin{itemize}
        \item The answer \answerNA{} means that the paper does not include experiments.
        \item The paper should indicate the type of compute workers CPU or GPU, internal cluster, or cloud provider, including relevant memory and storage.
        \item The paper should provide the amount of compute required for each of the individual experimental runs as well as estimate the total compute. 
        \item The paper should disclose whether the full research project required more compute than the experiments reported in the paper (e.g., preliminary or failed experiments that didn't make it into the paper). 
    \end{itemize}
    
\item {\bf Code of ethics}
    \item[] Question: Does the research conducted in the paper conform, in every respect, with the NeurIPS Code of Ethics \url{https://neurips.cc/public/EthicsGuidelines}?
    \item[] Answer: \answerYes{}
    \item[] Justification: We have reviewed the NeurIPS Code of Ethics and believe the research conforms to it. The submission is anonymized and discusses potential misuse risks of fast controllable generation.
    \item[] Guidelines:
    \begin{itemize}
        \item The answer \answerNA{} means that the authors have not reviewed the NeurIPS Code of Ethics.
        \item If the authors answer \answerNo, they should explain the special circumstances that require a deviation from the Code of Ethics.
        \item The authors should make sure to preserve anonymity (e.g., if there is a special consideration due to laws or regulations in their jurisdiction).
    \end{itemize}

\item {\bf Broader impacts}
    \item[] Question: Does the paper discuss both potential positive societal impacts and negative societal impacts of the work performed?
    \item[] Answer: \answerYes{}
    \item[] Justification: The paper discusses positive impacts such as efficient controllable generation, and negative impacts such as lowering the cost of generating misleading visual content. It also discusses responsible release practices such as usage restrictions, provenance tools, and safety filters.
    \item[] Guidelines:
    \begin{itemize}
        \item The answer \answerNA{} means that there is no societal impact of the work performed.
        \item If the authors answer \answerNA{} or \answerNo, they should explain why their work has no societal impact or why the paper does not address societal impact.
        \item Examples of negative societal impacts include potential malicious or unintended uses (e.g., disinformation, generating fake profiles, surveillance), fairness considerations (e.g., deployment of technologies that could make decisions that unfairly impact specific groups), privacy considerations, and security considerations.
        \item The conference expects that many papers will be foundational research and not tied to particular applications, let alone deployments. However, if there is a direct path to any negative applications, the authors should point it out. For example, it is legitimate to point out that an improvement in the quality of generative models could be used to generate Deepfakes for disinformation. On the other hand, it is not needed to point out that a generic algorithm for optimizing neural networks could enable people to train models that generate Deepfakes faster.
        \item The authors should consider possible harms that could arise when the technology is being used as intended and functioning correctly, harms that could arise when the technology is being used as intended but gives incorrect results, and harms following from (intentional or unintentional) misuse of the technology.
        \item If there are negative societal impacts, the authors could also discuss possible mitigation strategies (e.g., gated release of models, providing defenses in addition to attacks, mechanisms for monitoring misuse, mechanisms to monitor how a system learns from feedback over time, improving the efficiency and accessibility of ML).
    \end{itemize}
    
\item {\bf Safeguards}
    \item[] Question: Does the paper describe safeguards that have been put in place for responsible release of data or models that have a high risk for misuse (e.g., pre-trained language models, image generators, or scraped datasets)?
    \item[] Answer: \answerYes{}
    \item[] Justification: The paper discusses safeguards for responsible release, including usage restrictions, provenance or watermarking tools when possible, and content-safety filters inherited from the underlying pretrained generators.
    \item[] Guidelines:
    \begin{itemize}
        \item The answer \answerNA{} means that the paper poses no such risks.
        \item Released models that have a high risk for misuse or dual-use should be released with necessary safeguards to allow for controlled use of the model, for example by requiring that users adhere to usage guidelines or restrictions to access the model or implementing safety filters. 
        \item Datasets that have been scraped from the Internet could pose safety risks. The authors should describe how they avoided releasing unsafe images.
        \item We recognize that providing effective safeguards is challenging, and many papers do not require this, but we encourage authors to take this into account and make a best faith effort.
    \end{itemize}

\item {\bf Licenses for existing assets}
    \item[] Question: Are the creators or original owners of assets (e.g., code, data, models), used in the paper, properly credited and are the license and terms of use explicitly mentioned and properly respected?
    \item[] Answer: \answerYes{}
    \item[] Justification: The paper cites the datasets, pretrained models, and baseline methods used in the experiments. We use these assets consistently with their stated licenses and terms of use.
    \item[] Guidelines:
    \begin{itemize}
        \item The answer \answerNA{} means that the paper does not use existing assets.
        \item The authors should cite the original paper that produced the code package or dataset.
        \item The authors should state which version of the asset is used and, if possible, include a URL.
        \item The name of the license (e.g., CC-BY 4.0) should be included for each asset.
        \item For scraped data from a particular source (e.g., website), the copyright and terms of service of that source should be provided.
        \item If assets are released, the license, copyright information, and terms of use in the package should be provided. For popular datasets, \url{paperswithcode.com/datasets} has curated licenses for some datasets. Their licensing guide can help determine the license of a dataset.
        \item For existing datasets that are re-packaged, both the original license and the license of the derived asset (if it has changed) should be provided.
        \item If this information is not available online, the authors are encouraged to reach out to the asset's creators.
    \end{itemize}

\item {\bf New assets}
    \item[] Question: Are new assets introduced in the paper well documented and is the documentation provided alongside the assets?
    \item[] Answer: \answerNA{}
    \item[] Justification: The paper does not introduce a new dataset or benchmark. If code or checkpoints are released after publication, they will be documented with training details, intended use, limitations, and license information.
    \item[] Guidelines:
    \begin{itemize}
        \item The answer \answerNA{} means that the paper does not release new assets.
        \item Researchers should communicate the details of the dataset\slash code\slash model as part of their submissions via structured templates. This includes details about training, license, limitations, etc. 
        \item The paper should discuss whether and how consent was obtained from people whose asset is used.
        \item At submission time, remember to anonymize your assets (if applicable). You can either create an anonymized URL or include an anonymized zip file.
    \end{itemize}

\item {\bf Crowdsourcing and research with human subjects}
    \item[] Question: For crowdsourcing experiments and research with human subjects, does the paper include the full text of instructions given to participants and screenshots, if applicable, as well as details about compensation (if any)? 
    \item[] Answer: \answerNo{}
    \item[] Justification: The paper reports a user study, but the current version does not include the full participant instructions, screenshots, or compensation details. These details should be added to the appendix or supplemental material.
    \item[] Guidelines:
    \begin{itemize}
        \item The answer \answerNA{} means that the paper does not involve crowdsourcing nor research with human subjects.
        \item Including this information in the supplemental material is fine, but if the main contribution of the paper involves human subjects, then as much detail as possible should be included in the main paper. 
        \item According to the NeurIPS Code of Ethics, workers involved in data collection, curation, or other labor should be paid at least the minimum wage in the country of the data collector. 
    \end{itemize}

\item {\bf Institutional review board (IRB) approvals or equivalent for research with human subjects}
    \item[] Question: Does the paper describe potential risks incurred by study participants, whether such risks were disclosed to the subjects, and whether Institutional Review Board (IRB) approvals (or an equivalent approval/review based on the requirements of your country or institution) were obtained?
    \item[] Answer: \answerNo{}
    \item[] Justification: The paper includes a user study, but the current version does not state whether IRB approval, exemption, or equivalent review was obtained. This should be clarified in the appendix while preserving anonymity.
    \item[] Guidelines:
    \begin{itemize}
        \item The answer \answerNA{} means that the paper does not involve crowdsourcing nor research with human subjects.
        \item Depending on the country in which research is conducted, IRB approval (or equivalent) may be required for any human subjects research. If you obtained IRB approval, you should clearly state this in the paper. 
        \item We recognize that the procedures for this may vary significantly between institutions and locations, and we expect authors to adhere to the NeurIPS Code of Ethics and the guidelines for their institution. 
        \item For initial submissions, do not include any information that would break anonymity (if applicable), such as the institution conducting the review.
    \end{itemize}

\item {\bf Declaration of LLM usage}
    \item[] Question: Does the paper describe the usage of LLMs if it is an important, original, or non-standard component of the core methods in this research? Note that if the LLM is used only for writing, editing, or formatting purposes and does \emph{not} impact the core methodology, scientific rigor, or originality of the research, declaration is not required.
    %this research? 
    \item[] Answer: \answerNA{}
    \item[] Justification: The core method, experiments, and scientific claims do not use LLMs as an important, original, or non-standard component. Any writing or formatting assistance does not affect the methodology or results.
    \item[] Guidelines:
    \begin{itemize}
        \item The answer \answerNA{} means that the core method development in this research does not involve LLMs as any important, original, or non-standard components.
        \item Please refer to our LLM policy in the NeurIPS handbook for what should or should not be described.
    \end{itemize}

\end{enumerate}